\documentclass[journal]{IEEEtran}
\usepackage{xcolor,soul,framed} 
\pdfoutput=1
\colorlet{shadecolor}{yellow}
\usepackage[pdftex]{graphicx}

\usepackage[cmex10]{amsmath}
\usepackage{array}
\usepackage{mdwmath}
\usepackage{mdwtab}
\usepackage{eqparbox}
\usepackage{url}
\usepackage{cite}
\usepackage{graphicx}
\usepackage{amsmath}
\DeclareMathOperator*{\argmin}{min} 

\usepackage{amssymb}
\usepackage{amsthm}
\theoremstyle{definition}
\theoremstyle{remark}
\usepackage{multirow,tabularx}
\usepackage{mathtools}
\usepackage{units}
\usepackage{cite}
\usepackage{makecell}
\usepackage{etoolbox}
\newcolumntype{Y}{>{\centering\arraybackslash}X}
\usepackage{multirow}
\usepackage{booktabs}
\usepackage{slashbox}

\usepackage{silence}
\let\mybibitem\bibitem
\renewcommand{\bibitem}[1]{%
  \ifstrequal{#1}{Zhang_1}
    {\color{black}\mybibitem{#1}}
    {\color{black}\mybibitem{#1}}%
}

\usepackage{tikz}
\usetikzlibrary{arrows,positioning,shapes.geometric}
\usetikzlibrary{positioning,fit,calc}
\tikzset{block/.style={draw,thick,text width=0.5cm,minimum height=1cm,align=center},
         line/.style={-latex}
}
\usepackage{textcomp}
\usetikzlibrary{fit}
\usetikzlibrary{arrows,decorations.pathmorphing,backgrounds,fit,positioning,shapes.symbols,chains}

\usepackage{forest}

\usepackage{tikz}       
\usetikzlibrary{arrows,positioning,patterns,decorations.pathmorphing,calc}

\DeclareMathAlphabet{\mathpzc}{OT1}{pzc}{m}{it}

\usetikzlibrary{arrows,positioning,shapes.geometric}
\usetikzlibrary{positioning,fit,calc}
\tikzset{block/.style={draw,thick,text width=1cm,minimum height=1cm,align=center},
         line/.style={-latex},
         comment/.style={rectangle, inner sep= 1pt, text width=0.25cm, node distance=0.25cm,
         font=\scriptsize\sffamily},
}

\usetikzlibrary{arrows,positioning,patterns,decorations.pathmorphing,calc}
\tikzstyle{startstop} = [rectangle, rounded corners, minimum width=0.5cm, minimum height=0.25cm,text centered, draw=black]
\tikzstyle{arrow} = [thick,->,>=stealth]

\definecolor{hadi}{rgb}{0.33, 0.42, 0.18} 

\usepackage{scalerel}
\usepackage{tikz}
\usetikzlibrary{svg.path}

\definecolor{orcidlogocol}{HTML}{A6CE39}
\tikzset{
  orcidlogo/.pic={
    \fill[orcidlogocol] svg{M256,128c0,70.7-57.3,128-128,128C57.3,256,0,198.7,0,128C0,57.3,57.3,0,128,0C198.7,0,256,57.3,256,128z};
    \fill[white] svg{M86.3,186.2H70.9V79.1h15.4v48.4V186.2z}
                 svg{M108.9,79.1h41.6c39.6,0,57,28.3,57,53.6c0,27.5-21.5,53.6-56.8,53.6h-41.8V79.1z M124.3,172.4h24.5c34.9,0,42.9-26.5,42.9-39.7c0-21.5-13.7-39.7-43.7-39.7h-23.7V172.4z}
                 svg{M88.7,56.8c0,5.5-4.5,10.1-10.1,10.1c-5.6,0-10.1-4.6-10.1-10.1c0-5.6,4.5-10.1,10.1-10.1C84.2,46.7,88.7,51.3,88.7,56.8z};
  }
}

\newcommand\orcidicon[1]{\href{https://orcid.org/#1}{\mbox{\scalerel*{
\begin{tikzpicture}[yscale=-1,transform shape]
\pic{orcidlogo};
\end{tikzpicture}
}{|}}}}

\usepackage{hyperref} 

\begin{document}

    \title{Generating Black-Box Adversarial Examples in Sparse Domain}
  \author{Hadi Zanddizari \orcidicon{0000-0002-2465-9374},~\IEEEmembership{Student Member,~IEEE,}
      Behnam Zeinali \orcidicon{0000-0003-3916-5841},~\IEEEmembership{Student Member,~IEEE,}\\
      J. Morris Chang \orcidicon{0000-0002-0660-7191},~\IEEEmembership{Senior Member,~IEEE}

  \thanks{Hadi Zanddizari, Behnam Zeinali, and J. Morris Chang are with the Department of Electrical Engineering, University of South Florida, Tampa, Florida, USA. (e-mail: hadiz@usf.edu; behnamz@usf.edu; chang5@usf.edu).}

 \thanks{\textcopyright 2021 IEEE. Personal use of this material is permitted.  Permission from IEEE must be obtained for all other uses, in any current or future media, including reprinting/republishing this material for advertising or promotional purposes, creating new collective works, for resale or redistribution to servers or lists, or reuse of any copyrighted component of this work in other works.}}

\markboth{This Paper has been accepted in IEEE Transactions on Emerging Topics in Computational Intelligence}{}


\maketitle
\begin{abstract}
Applications of machine learning (ML) models and convolutional neural networks (CNNs) have been rapidly increased. Although state-of-the-art CNNs provide high accuracy in many applications, recent investigations show that such networks are highly vulnerable to adversarial attacks. The black-box adversarial attack is one type of attack that the attacker does not have any knowledge about the model or the training dataset, but it has some input data set and their labels. 

In this paper, we propose a novel approach to generate a black-box attack in sparse domain whereas the most important information of an image can be observed. Our investigation shows that large sparse (LaS) components play a critical role in the performance of image classifiers. Under this presumption, to generate adversarial example, we transfer an image into a sparse domain and put a threshold to choose only $k$ LaS components. In contrast to the very recent works that randomly perturb $k$ low frequency (LoF) components, we perturb $k$ LaS components either randomly (query-based) or in the direction of the most correlated sparse signal from a different class. We show that LaS components contain some middle or higher frequency components information which leads fooling image classifiers with a fewer number of queries. 
We demonstrate the effectiveness of this approach by fooling six state-of-the-art image classifiers, the TensorFlow Lite (TFLite) model of Google Cloud Vision platform, and YOLOv5 model as an object detection algorithm. Mean squared error (MSE) and peak signal to noise ratio (PSNR) are used as quality metrics. We also present a theoretical proof to connect these metrics to the level of perturbation in the sparse domain. \end{abstract}

\begin{IEEEkeywords}
Convolutional neural network, black-box attack, deep learning, sparse representation
\end{IEEEkeywords}

\IEEEpeerreviewmaketitle

\section{Introduction} \label{intro}
\IEEEPARstart{B}{y} the ever-increasing demands for analyzing and processing large datasets, ML algorithms and particularly deep learning techniques have become the center of attention of many companies and service providers. The remarkable performance of CNNs for image segmentation, classification, and object tracking could provide acceptable solutions for many problems encountered in computer vision and biomedical engineering.   \cite{LeCun,demotalogist_CNN, hb-skin}. While almost CNNs perform well and provide high accuracy, their robustness toward some malicious attacks still are not acceptable\cite{Papernot1, Szegedy1, vulnerability_CNN}. Applying some perturbation on the input data may totally undermine the high accuracy of a classifier since ML models are usually trained and deployed in benign settings.  In other words, they do not consider certain scenarios in which an attacker can compromise the performance of the system. 

Recently, many works have been proposed to point out the vulnerability of CNNs against adversarial scenarios \cite{threat_CNN1, threat_CNN2, threat_CNN3, threat_CNN4, threat_CNN5}. By slightly perturbing the input data, ML classifier may fool and predict a wrong label. If this perturbation is small enough to the human eyes, then the perturbed image is called an adversarial example \cite{Szegedy1,survery2018,survery2017}. This problem can be viewed from a different perspective, if we add a limited perturbation to an image, while human eyes may detect the perturbation, but still we expect the classifiers classify correctly. It opens up a new horizon of the robustness of ML models against adversarial examples. 


An adversarial example can be obtained by solving the following minimization problem
\begin{equation}
\argmin ||\mathbf{r}||_2 \quad  \text{s.t.} \quad C(\mathbf{x}+\mathbf{r}) \neq C(\mathbf{x})
\label{eq:l1}
\end{equation}
where $\mathbf{r}$ is adversarial perturbation, $||.||_2$ is the Euclidean norm or $\ell_{2}$ norm, $\mathbf{x}$ is the legitimate image (original image), and $C(.)$ yields the classifier's output label. Based on (\ref{eq:l1}), there are two factors in generating adversarial examples, first having a minimum perturbation on the legitimate image, and the second, fooling the classifier output.  


Misclassification and targeted misclassification attacks are two major goals of adversarial examples. In the misclassification attack, an adversary tries to fool the ML classifier by misclassifying a legitimate example to different classes other than the original one. For example, a legitimate image with a label \lq$1$\rq of the MNIST (Modified National Institute of Standards and Technology) dataset is perturbed in such a way that ML classifier yields an output label belongs to $\{0,2,3,4,5,6,7,8,9\}$, yet not \lq$1$\rq. In targeted misclassification, the attacker tries to fool the classifier to yield a targeted label. For example, the same legitimate image with a label \lq$1$\rq is labeled as a specific number like \lq$8$\rq   by the classifier. In this study, we focus on misclassification attacks.  

Adversarial examples can be generated based on two different approaches: white-box and black-box. In white-box attacks, the attacker has comprehensive knowledge about the training dataset, model's parameters, number of CNN layers, loss function, and the whole structure of the model. There are numerous works based on white-box attacks, such as fast gradient sign method (FGSM) \cite{FGSM}, beyond the image space approach that uses physical space features of 3D images, \cite{Adv_beyondImgSpace}, deepfool \cite{DeepFool}, Jacobean-based Saliency Map Attack (JSMA)\cite{JSMA}. For example, FGSM generates an adversarial perturbation for a given legitimate image by computing the gradient of the cost function with respect to the legitimate image of the ML algorithm as follows:

\begin{equation}
\mathbf{x}^\ast=\mathbf{x}+\epsilon\ sign\left(\nabla_\mathbf{x}\mathcal{J}\left(\mathbf{x},c\right)\right)    \label{eq:l2}
\end{equation}
where  $\epsilon$ denotes a small scalar value which regulates the perturbation's level, $c$ is the input label, $\mathcal{J}()$ denotes the model cost function, $\nabla_x$  is the gradient of the trained model with respect to the legitimate image,  and $sign(.)$ is the common mathematical function which yields the sign of its input argument. The common property of white-box attacks is utilizing the model's information for generating the adversarial example. In contrast, the black-box attack does not have any information about the model's structure and parameters, and training dataset\cite{Papernot2017,Transferability,black1,black2}. This type of attack is more practical because in many cases having access to the training dataset is not possible. Also, some information such as the model's parameters, number of layers, and loss function may not be public. 

Black-box attacks can be separated into three categories: non-adaptive, adaptive and strictly black-box attacks \cite{survery2018}. 
In a non-adaptive black-box attack, an attacker can have access only to the distribution of the training dataset \cite{non_adapt_adv3}. In the adaptive black-box case, the attacker does not have any information about the distribution of the dataset, however she can access the target model as an oracle. It means, the attacker can query the output labels of legitimate samples as well as adversarial samples \cite{Low_DCT2, Low_DCT1}. In the strict black-box attack, the attacker does not have access to the training distribution of the dataset and also she cannot adaptively modify the input query to observe the model's output. In other words, an attacker can query the legitimate input samples, but if she slightly perturbs an input sample to observe its output label, the system identifies this process as a malicious attack \cite{strict_adv, survery2018}. Although these types of systems may provide high level of security, in many real cases input samples may be very similar to each other and as a result, there is no need to block the user. Adaptive black-box attacks are more applicable than non-adaptive or strict black-box attacks as they do not have any knowledge about the distribution of the training dataset and assumes the system would not block a user by evaluating a limited number of close queries. However, if the number of queries increases, the system may detect a probable malicious attack.\\
In \cite{onePixel_adv}, authors proposed generating adversarial examples based on perturbing one-pixel of an image through differential evolution. Although this method could fool almost CNN models due to  the inherent features of differential evolution, there is no limit for the number of queries to attack the model. Papernot et al. \cite{Papernot2017} proposed a practical approach for generating adversarial examples based on Jacobian-based dataset augmentation technique to obtain new synthetic training samples. After having an adequate number of samples and corresponding labels, they train a local model and apply a white-box attack (such as FGSM) on this locally trained model to generate adversarial examples. They use the transferability property of ML algorithms \cite{Transferability}. Transferability is a property that enables us to apply adversarial examples generated by a model on another model with the same or different architecture. The applicability of such attacks mainly revolves around the transferability property of ML models and having enough large dataset for training the local model. Recently, Hosseini et al. \cite{NullLabeling} proposed a three-step null labeling method to block the transferability property of the ML models. In the first step, they train the model based on clean data, then they add some perturbations to the input data, and based on some threshold and probability functions, they assign the label \lq $Null$\rq\ to the perturbed image. Then, they retrain the model with clean and new adversarial examples which have null labels. This approach enables the model to detect the input adversarial examples by predicting as a \lq $Null$\rq. 
The previous black-box attacks try to generate adversarial examples based on a white-box approach. In other words, they train a local fake model, then apply a white-box attack to generate adversarial examples.\\
There are some black-box approaches that are not based on the white-box approaches. In \cite{Low_DCT2}, the effectiveness of restricting the search for adversarial images to a low frequency domain has been investigated. After focusing on the lower frequency subspace, they randomly perturb the components while restricting the perturbation level. It can be described as adding a low-filtered random noise to the legitimate image. This approach could outperform many black-box attacks. Y. Sharma et al. \cite{Low_DCT1} used discrete cosine transform (DCT) dictionary to map the image into the frequency domain, then they put a hard threshold for choosing LoF components. After transformation into the frequency domain, most of the frequency components have small values and only a few of them have large values. This property of the frequency domain is well known as a sparse representation of an image. Then, by applying perturbations on the LoF components, they could generate faster and more transferable adversarial examples. This approach can completely bypass most of the top-placing defense strategies at the NeurIPS 2017 competition. The authors also investigated the effect of perturbation on high frequency (HiF) components, but their results show that LoF components are the ones that mostly affect CNN models. We motivated by the aforementioned work and used DCT dictionary to transfer images into the sparse (frequency) domain. Then, instead of putting a hard threshold for choosing only $k$ LoF components, we selected $k$ LaS components where some low, middle, and high frequency components are picked up. In section \ref{laSvsLoF}, we show the difference between LaS and LoF components.\\
Focusing on LaS components have been used in many image processing and compression techniques. The JPEG codec \cite{JPEG} takes advantage of this property in order to compress the images. Because, the most critical features and information of an image are available in the LaS components and not just LoF components \cite{JPEG}. Intuitively, image classifiers are mostly consider specific components which bear more information of an image. We verify this property of image classifiers by implementing systematic experiments (section \ref{effectOfSparsity}). We propose adding noise to LaS components in two scenarios. In the first scenario, we randomly perturb LaS components, and by restricting the perturbation level, the number of required queries to fool the state-of-the-art classifiers are evaluated. Our experiment results show that the proposed approach can fool the classifiers with less number of queries compared to the very recent approach which works based on LoF components\cite{Low_DCT1}. In the second scenario, a directed attack, we suppose a few number of images from each class are available.  Given a legitimate image, we perturb its LaS components in the direction of the most correlated sparse sample from a different class. Our experiments show that this method can successfully fool the state-of-the-art CNN classifiers.

 In this study, the summary of our contributions are as follow:

\begin{itemize}
  \item We introduce a black-box approach to generate adversarial examples in the sparse domain in order to fool the ML algorithms such as CNN models, support vector machine (SVM) classifiers, object detection algorithm (YOLOv5), and model trained by the Google Cloud Vision API.
  
  \item In contrast to the very recent black-box attacks which focused on LoF components, we show that the LaS components can fool the classifiers with a fewer number of queries.
  
  \item We proposed an analytical approach to show the relation between the perturbation level in the sparse domain and its effect on the pixel domain. Our results show the proposed method decreases the number of required queries to fool the ML models and increases the misclassification rate of ML models.  

\end{itemize}

\section{Sparsity}

\begin{figure}[!ht]
    \centering
    \includegraphics[height = 2.8in, width = 3.4in]{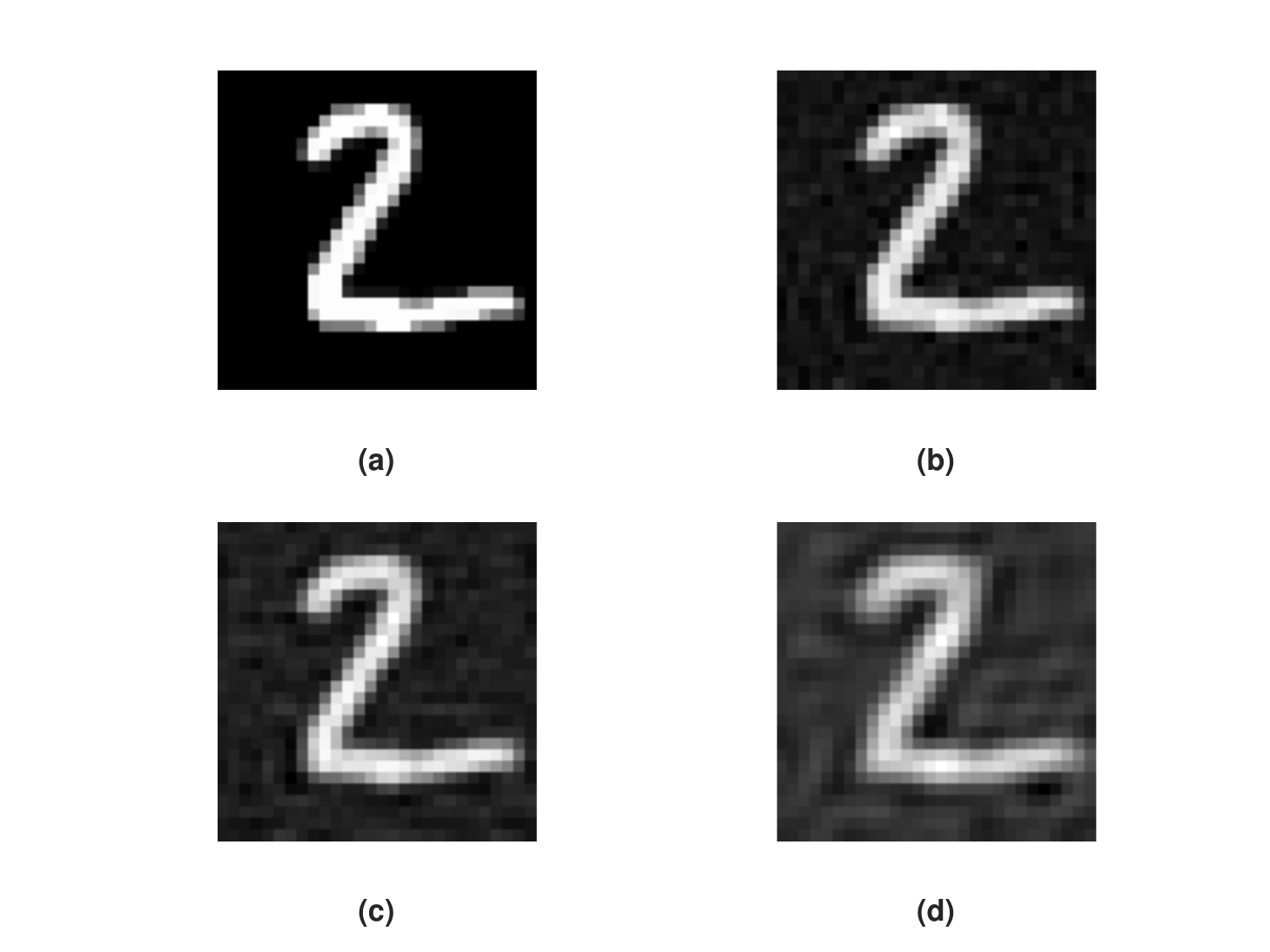}
    \caption{Transferring image into the sparse domain and zeroing small elements of sparse signal: (a) original image, (b) zeroing $70\%$, (c) $80\%$, and (d) $90\%$ of small elements. }
    \label{fig:fourimages}
\end{figure}

Sparsity has been widely used in many applications such as image denoising, deblurring, super resolution, and compression\cite{Img_Super,Img_denoising,ujan,Img_compress,CS1}.
 An image signal $\mathbf{X} \in \mathbb{R}^{p \times q}$ can be reshaped to a vector $\mathbf{x} \in \mathbb{R}^{N =p \times q}$
 where $N$ is the number of pixels. Dictionary $\mathbf{D} \in \mathbb{R}^{N \times L}$ is a matrix which linear combination of its columns $\mathbf{d}_i$ can approximately represent the $\mathbf{x}$  as follow

\begin{equation}
\mathbf{x}= \sum_{i\in\left\{1,2,..L\right\}}{s_i\mathbf{\mathbf{d}}_i}=\mathbf{D}\mathbf{s}
\label{eq:l3}
\end{equation}
where $\mathbf{s}\in\mathbb{R}^L$ is the weight vector. If $\mathbf{D}$ provides a weight vector with only $k$ large and $l-k$ negligible or zero elements, then $\mathbf{D}$ and $\mathbf{s}$ can be called as a sparsifying dictionary and sparse representation of input $\mathbf{x}$, respectively. For brevity, by the rest of this work, we omit the ‘sparsifying’ and refer to the ``dictionary'' as a sparsifying dictionary. There are some fixed dictionaries based on analytical approaches such as Fourier or wavelet transform which can be designed very fast. In this work, we used DCT dictionary which is an orthonormal matrix $(\mathbf{D}\in\mathbb{R}^{N\times N}\ and\ {||\mathbf{d_i ||}}_2 = 1)$. The coefficients of DCT dictionary can be obtained as follows,

\begin{equation}
\begin{split}
d_{i,j} = a_{i,j}\cos{\frac{\pi(2i-1)(j-1)}{2N}}~~~~~~~i,j\in{1,2,\dots,N}\\
a_{i,j} = 
\begin{cases}
  \sqrt{\frac{1}{N}}~~j=1 \\    
  \sqrt{\frac{2}{N}}~~j\neq1    
\end{cases}
~~~~~~~~~~~~~~~~~~~~~~~~~~~~~~~~~~~~
\end{split}
\label{eq:dct}
\end{equation}
where $d_{i,j}$ corresponds to the entry of $i$th row and $j$th column of DCT dictionary.  If we transfer an image into the DCT domain, zeroing small components will have negligible effects on the visual information of the image. For example, Fig. \ref{fig:fourimages} illustrates this property. The original image was transferred into the sparse domain via DCT dictionary and forced $70\%$, $80\%$, and $90\%$ of its small components to zero, then transformed back into the pixel domain. It is evident that reconstructed images based on only $30\%$, $20\%$, or $10\%$ of its LaS components can still preserve lots of visual information of the image.

\subsection{Difference  between LaS and LoF components}\label{laSvsLoF}

\begin{figure}[!ht]
    \centering
    \includegraphics[height = 2.9in, width = 3.4in]{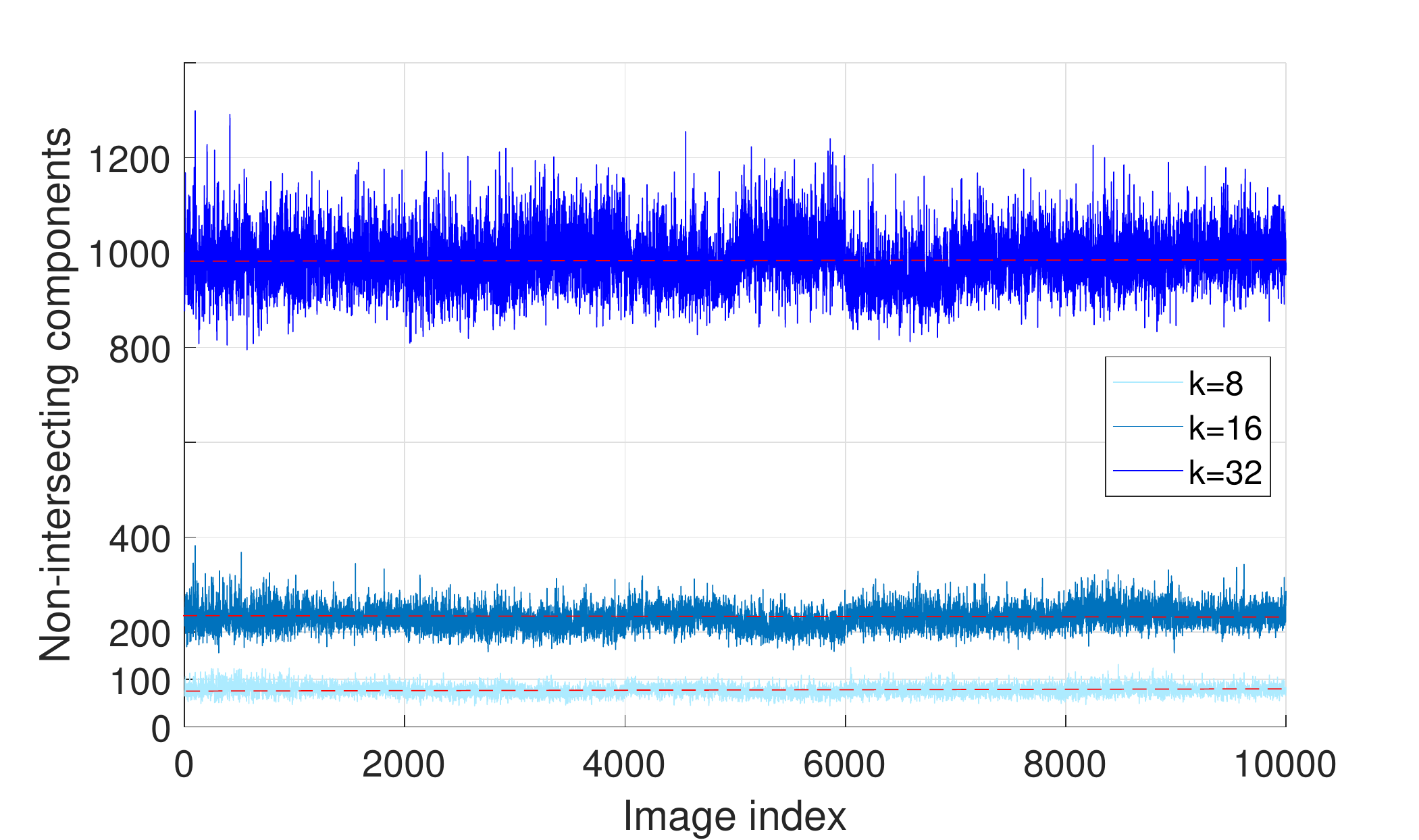}
    \caption{The number of non-intersecting components of each image. }
    \label{fig:intersection}
\end{figure}

Sparse domain enables us to have access to the important frequency components of an image. Components may belong to low, middle, or high frequency bands. Regardless of the frequency bands, if we choose some top-ranked components, those specific components can belong to any frequency bands. Some images may have some information in the middle or even higher frequencies, as a result, they would have LaS components corresponding to the middle or higher frequencies. To evaluate the level of intersection between LaS and LoF components, we used $10,000$ color images of size 256x256 pixels. The images had three color channels, and we mapped each channel into the sparse domain, separately. Then we selected $N = (k\times k\times 3)$ LaS and LoF components. For chosen $k = 8$, $k=16$, and $k=32$, the number of  components are $N=192$, $N=768$,  $N=3072$, respectively. Figure \ref{fig:intersection} shows how many non-intersecting components are available between LaS and LoF components. For $k=8$, the mean of non-intersecting components is $77$, i.e. more than $40\%$ of the LaS components belong to the middle or higher frequencies components. For  $k=16$ and  $k=32$ the mean of non-intersecting components are $229$ and $983$, i.e. $39\%$ and $32\%$ of the LaS components do not belong to the low frequency space. This experiment shows that the LaS components does not completely overlap with the LoF components, and some critical information of the image signals may belong to the middle or high frequency bands. In other words,for every image, different bands have different information, as a result, we cannot limit critical information of an image to only its low frequency space. In next section, we evaluate the effects of manipulating different frequency bands on the performance of CNN models. 

\begin{table*}[]
\centering
\label{table1}
\caption{The effect of keeping only 50\% or 30\% of LaS, LoF, and HiF components on the accuracy of six CNN models(\%).}
\begin{tabular}{|l|c|c|c|c|c|c|c|}
\hline
\multicolumn{1}{|c|}{\multirow{2}{*}{Model}} & \multirow{2}{*}{\begin{tabular}[c]{@{}c@{}}Ground Truth Accuracy \\(All components)\end{tabular}} & \multirow{2}{*}{\textbf{\begin{tabular}[c]{@{}c@{}}50\% of \\ LaS \end{tabular}}} & \multirow{2}{*}{\textbf{\begin{tabular}[c]{@{}c@{}}30\% of \\ LaS \end{tabular}}} & \multirow{2}{*}{\textbf{\begin{tabular}[c]{@{}c@{}}50\% of \\ LoF \cite{Low_DCT2} \& \cite{Low_DCT1} \end{tabular}}} & \multirow{2}{*}{\textbf{\begin{tabular}[c]{@{}c@{}}30\% of \\ LoF \cite{Low_DCT2} \& \cite{Low_DCT1} \end{tabular}}} & \multirow{2}{*}{\textbf{\begin{tabular}[c]{@{}c@{}}50\% of\\ HiF \cite{Low_DCT1}  \end{tabular}}} & \multirow{2}{*}{\textbf{\begin{tabular}[c]{@{}c@{}}30\% of \\ HiF \cite{Low_DCT1} \end{tabular}}} \\
\multicolumn{1}{|c|}{}                       &                                                                                                         &                                                                                              &                                                                                               &                                                                                               &                                                                                               &                                                                                              &                                                                                            
                                                                                        \\ \hline
MobileNets                                    & 90.72                                                                                                   & 89.14                                                                                        & 83.75                                                                                         & 77.14                                                                                         & 76.27                                                                                         & 29.79                                                                                        & 15.71                                                                                         \\ \hline
ResNet50                                     & 91.37                                                                                                   & 90.73                                                                                        & 87.59                                                                                         & 79.30                                                                                         & 73.29                                                                                         & 20.89                                                                                        & 16.13                                                                                         \\ \hline
DenseNet121                                  & 92.29                                                                                                   & 91.27                                                                                        & 88.05                                                                                         & 79.76                                                                                         & 77.84                                                                                         & 26.31                                                                                        & 16.74                                                                                         \\ \hline
InceptionV3                                  & 93.27                                                                                                   & 92.6                                                                                         & 90.32                                                                                         & 80.83                                                                                         & 79.40                                                                                         & 31.42                                                                                        & 25.93                                                                                         \\ \hline
Efficient-B0                                & 94.30                                                                                                   & 93.83                                                                                        & 90.59                                                                                         & 79.07                                                                                         & 70.57                                                                                         & 36.54                                                                                        & 27.16                                                                                         \\ \hline
Efficient-B1                                 & 95.46                                                                                                   & 94.78                                                                                        & 91.06                                                                                         & 80.25                                                                                         & 75.85                                                                                         & 37.36                                                                                        & 29.47                                                                                         \\ \hline
\end{tabular}
\end{table*}

\subsection{Effect of LaS components on CNN models}\label{effectOfSparsity}
Sparse transformation enables us to compact the energy of the signal into a few components. On the other hand, many image classifiers work based on pixel domain and they do not directly consider the sparse domain. A question that may arise here is: ``how much manipulating LaS, LoF, or HiF components can affect classifiers' performance?''. In this study, we empirically show that the LaS components are the most important part of images that affect the classifiers' performance. Our experiment was implemented over six state-of-the-art CNN models namely, EfficientNet-B0 and B1 \cite{EfficientNet}, ResNet50 \cite{Res_Net}, InceptionV3 \cite{InceptionV3}, MobileNets \cite{mobilenets}, and DenseNet121 \cite{DenseNet}. We used CIFAR-10 dataset which is a color and balanced image dataset with complex background. This dataset contains $50,000$ training samples and $10,000$ test samples belong to 10 classes. We trained these models with $50,000$ training samples, and then we input the original $10,000$ test samples (without any changes or manipulation) to obtain the ground truth accuracy of each trained model (Table I). In next step, via DCT dictionary we transferred all $10,000$ test samples into the sparse domain. Then we kept $50$\% and $30$\% of LaS, LoF, and HiF components, and zeroed the rest of the components. We transformed back each image to the pixel domain, and input them to the same trained model. To further clarify, after putting these thresholds, we obtained 6 test datasets, two for Las components, two for LoF components and two for HiF components.

As shown in Table I, the accuracies belong to LaS components test datasets are much closer to their corresponding ground truth accuracies. While  keeping only LoF or HiF components lead to considerable lost of accuracy. It shows that if we only focus on LoF or HiF components, we lose some components that affect the decision boundaries of CNN models. For example, Efficient-B1 which is one of the best image classifiers that has been introduced by Google in 2019, has the accuracy of $95.46$\% for the original test dataset. If we keep only $50$\% of LaS components, the accuracy is almost the same $94.78$\%. If we keep $50$\% of LoF and HiF components, the accuracies are $80.25$\% and $37.36$\%, respectively. To elucidate on, only $50$\% of LaS components affect classifiers, the other $50$\% components does not much affect the accuracy.  This experiment helps us to find out which frequency components mostly affect the CNN models. By having this information, we would be able to add perturbation on important components in order to fool image classifiers. Also this experiment verified the results of \cite{Low_DCT1} that showed the importance of LoF vs HiF components. They reached to this conclusion that perturbing LoF components is more effective than perturbing HiF components.
For the brevity, we omitted the results of our experiments over other CNN models, and different threshold levels which had the same results to verify our assumption. We release our code publicly for reproducibility. In next section, we add a limited perturbation to LaS and LoF components, to see which of them can fool the classifiers in a fewer number of queries.


\section{Perturbing LaS Components}
\label{PertLaSComponents}

\begin{figure*}[h]

\centering
    \includegraphics[height = 1.95in, width=0.47\linewidth]{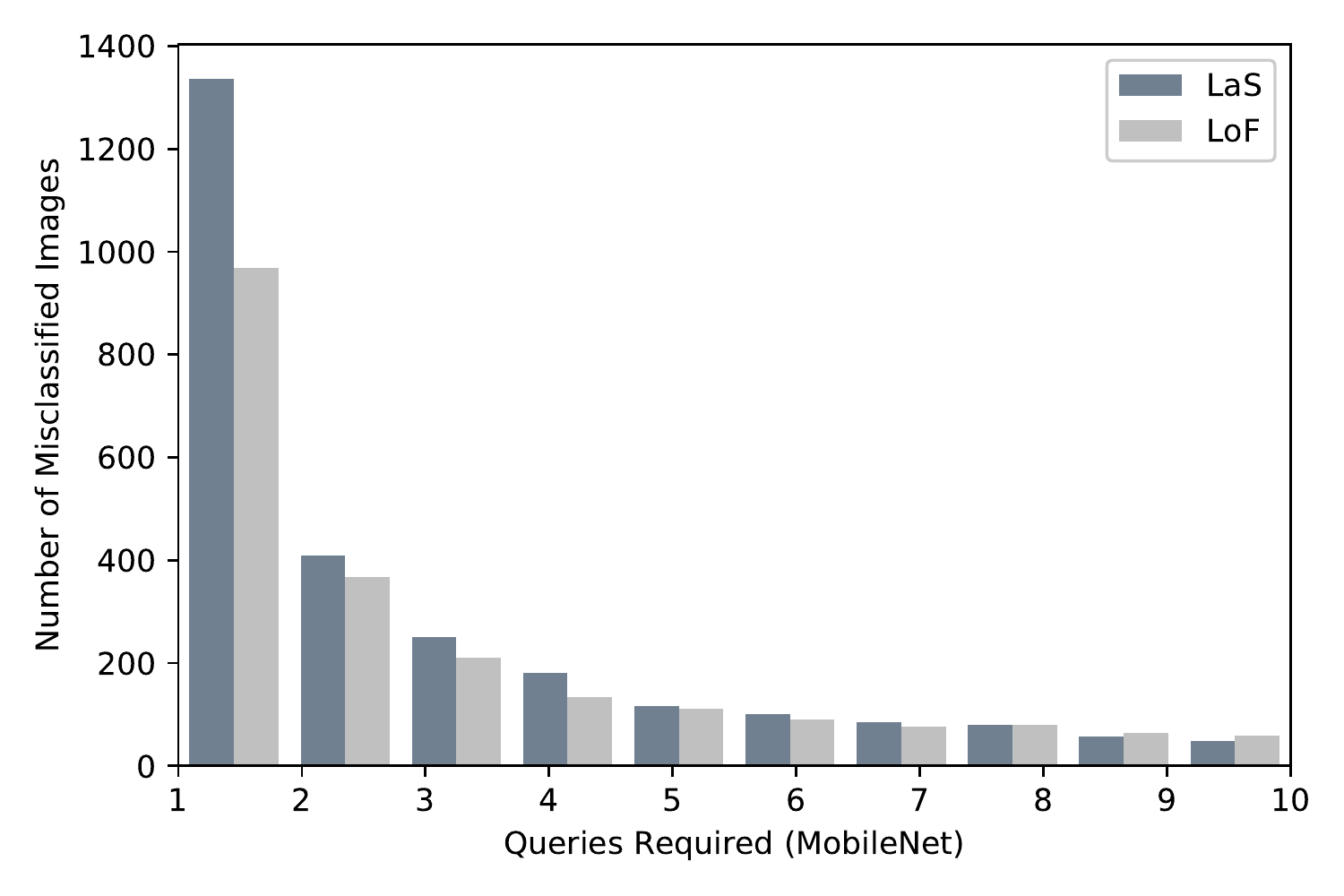}\hfil
    \includegraphics[height = 1.95in,width=0.47\linewidth]{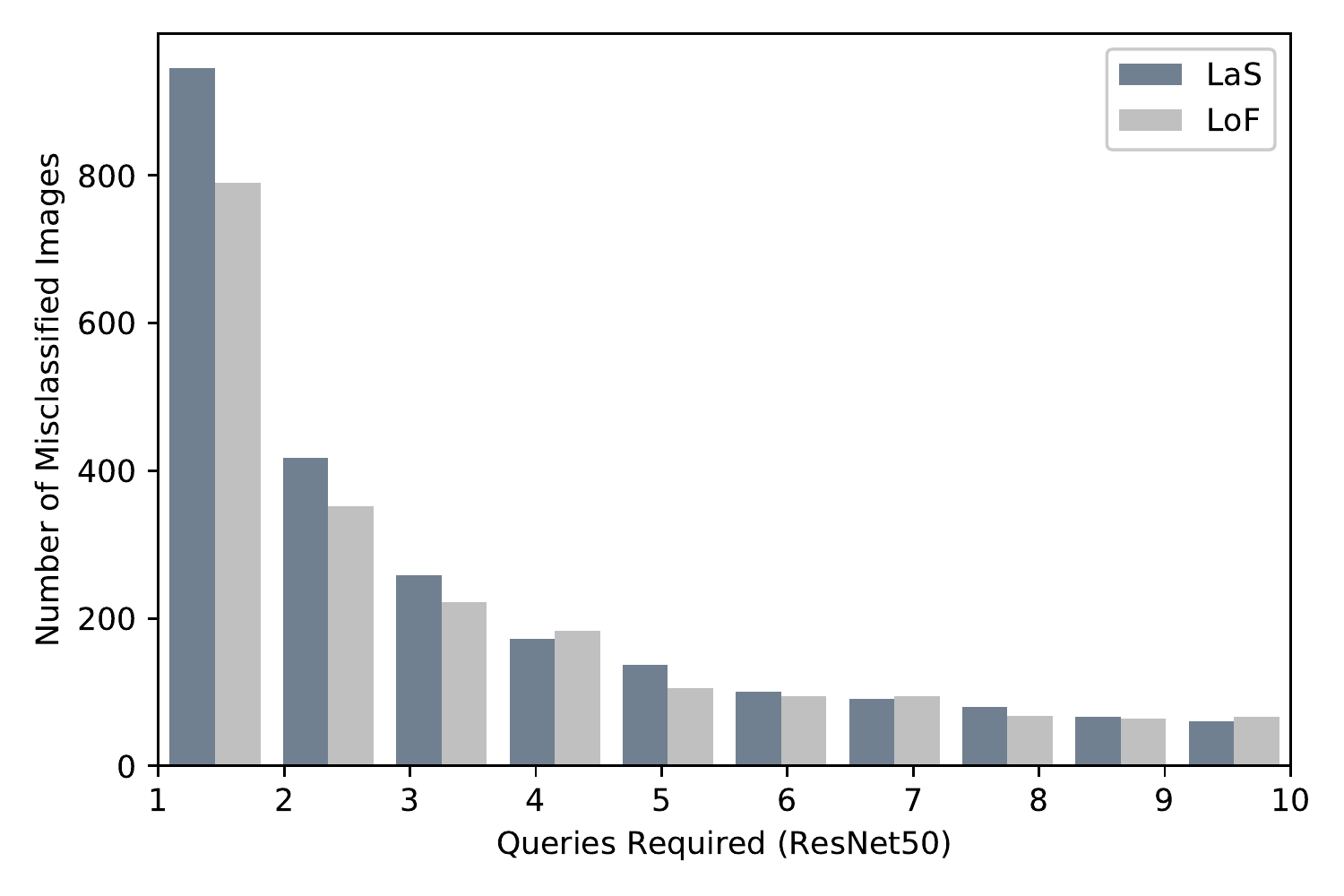}\par\medskip
    \includegraphics[height = 1.95in,width=0.47\linewidth]{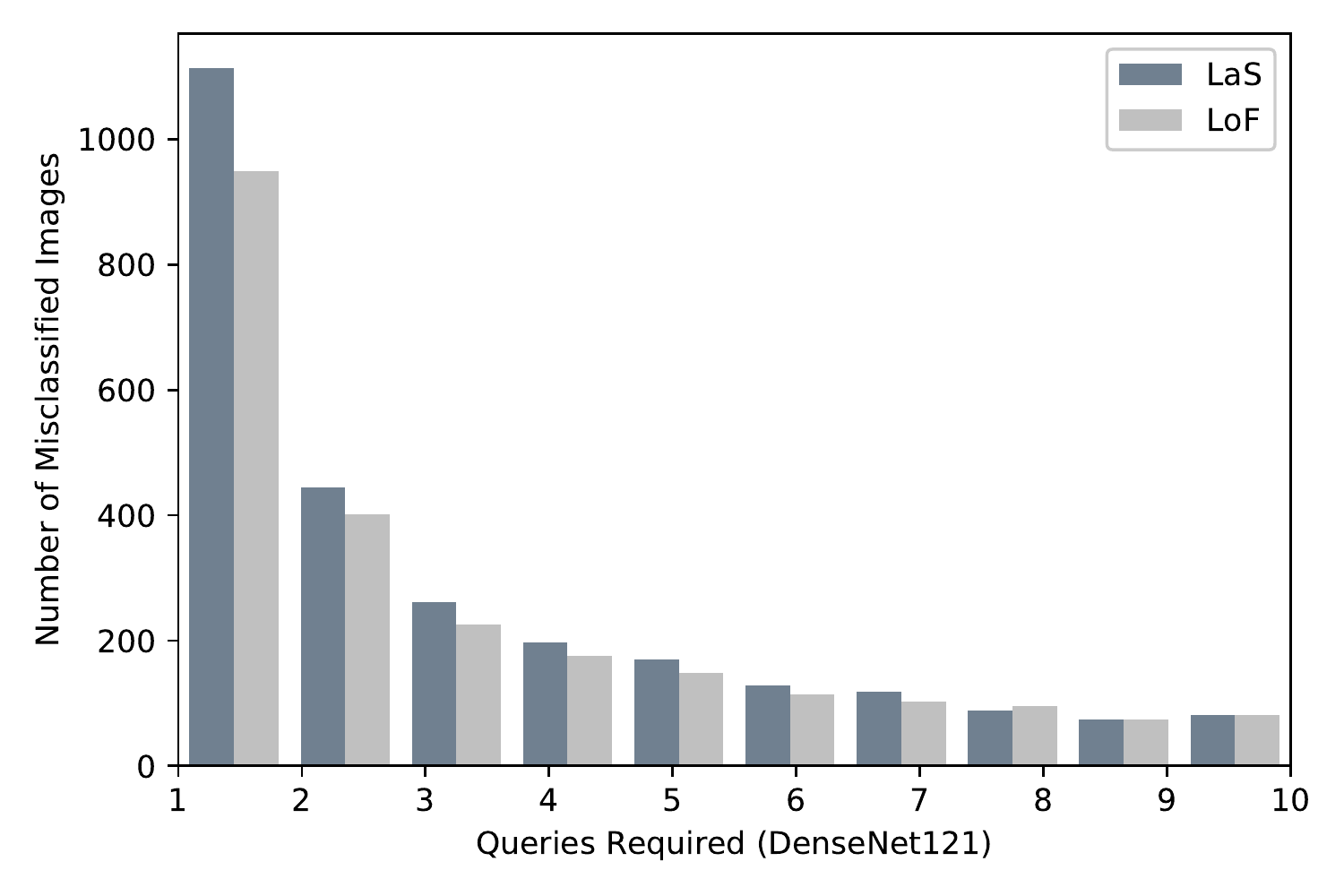}\hfil
    \includegraphics[height = 1.95in,width=0.47\linewidth]{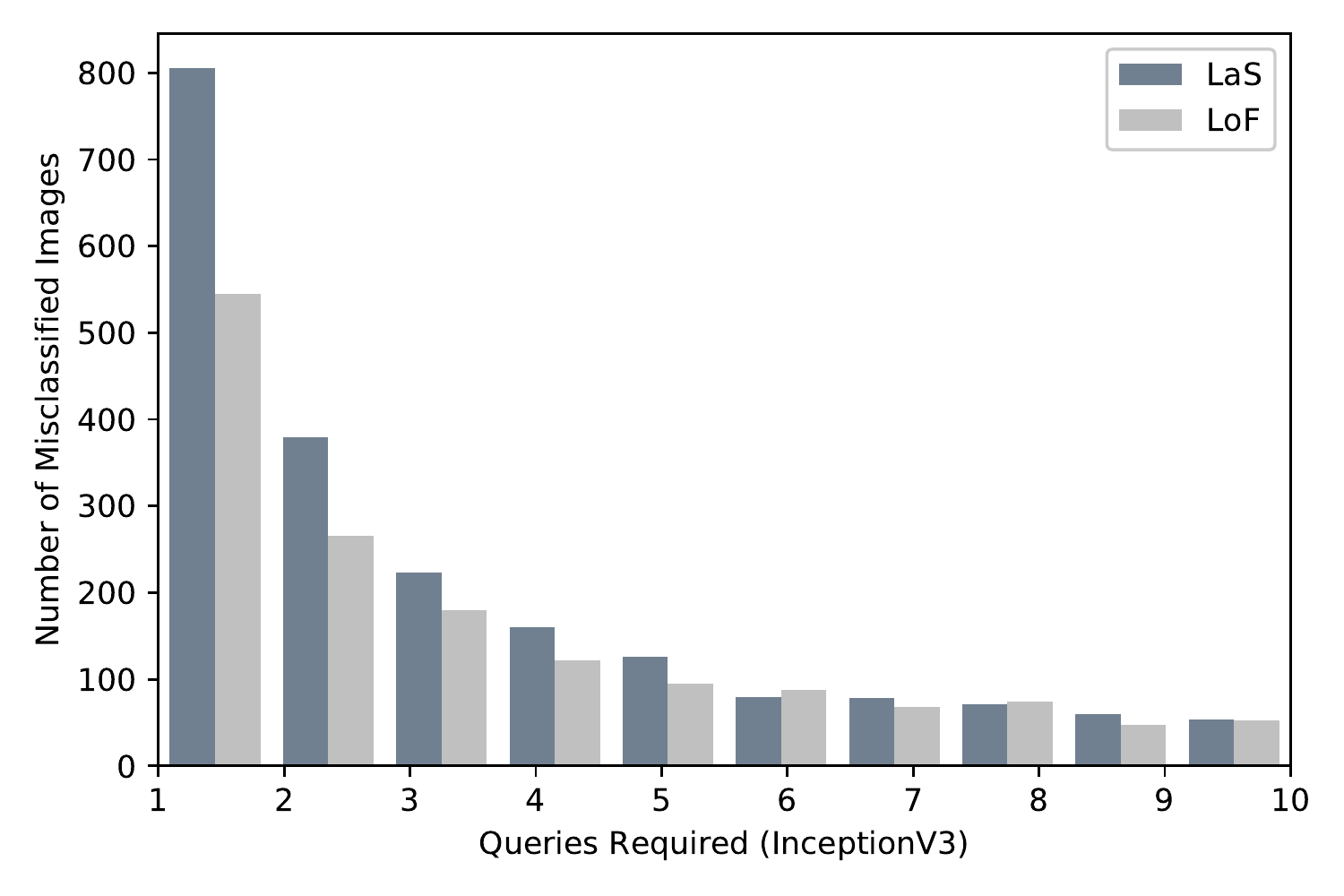}
    \includegraphics[height = 1.95in,width=0.47\linewidth]{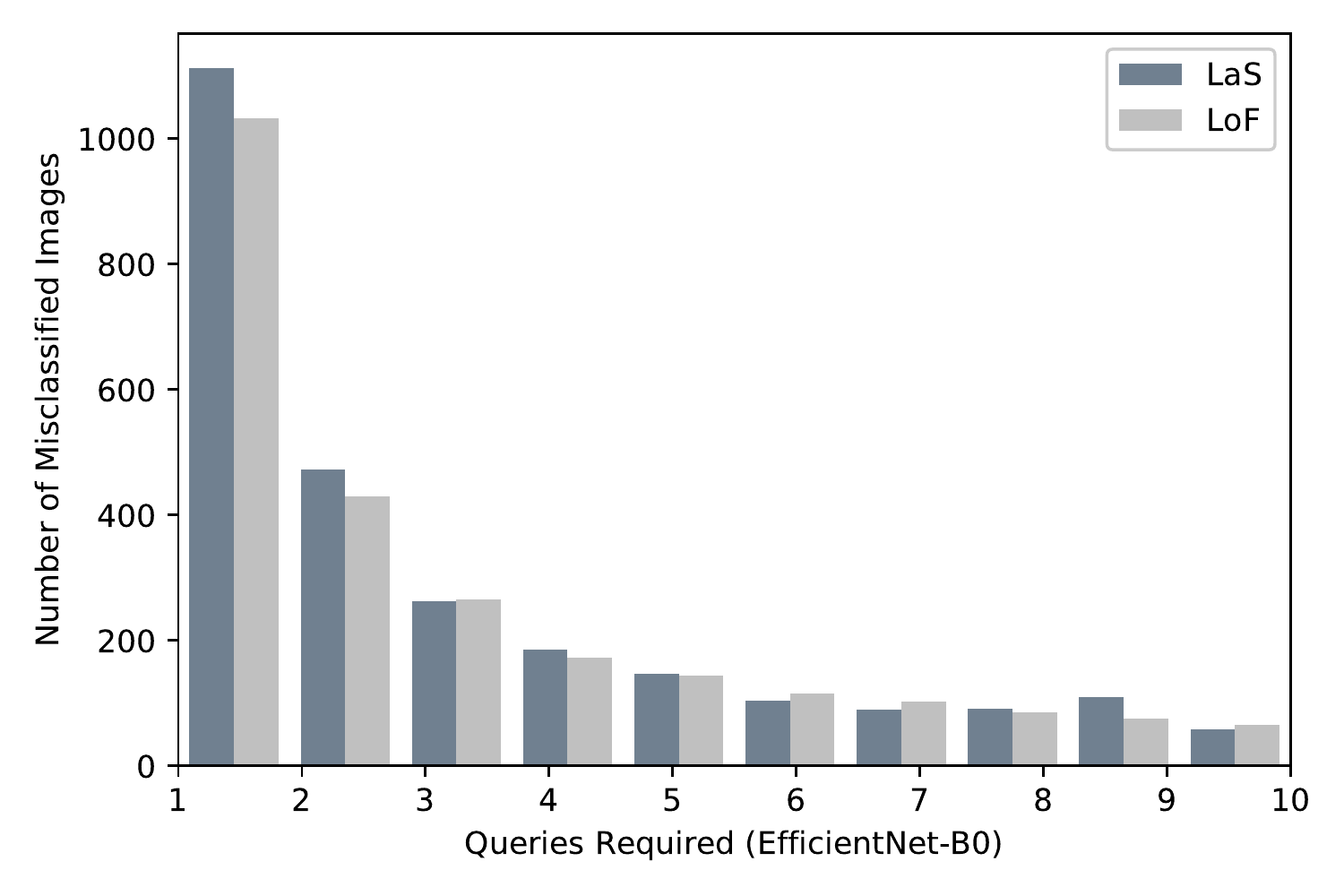}\hfil
    \includegraphics[height = 1.95in,width=0.47\linewidth]{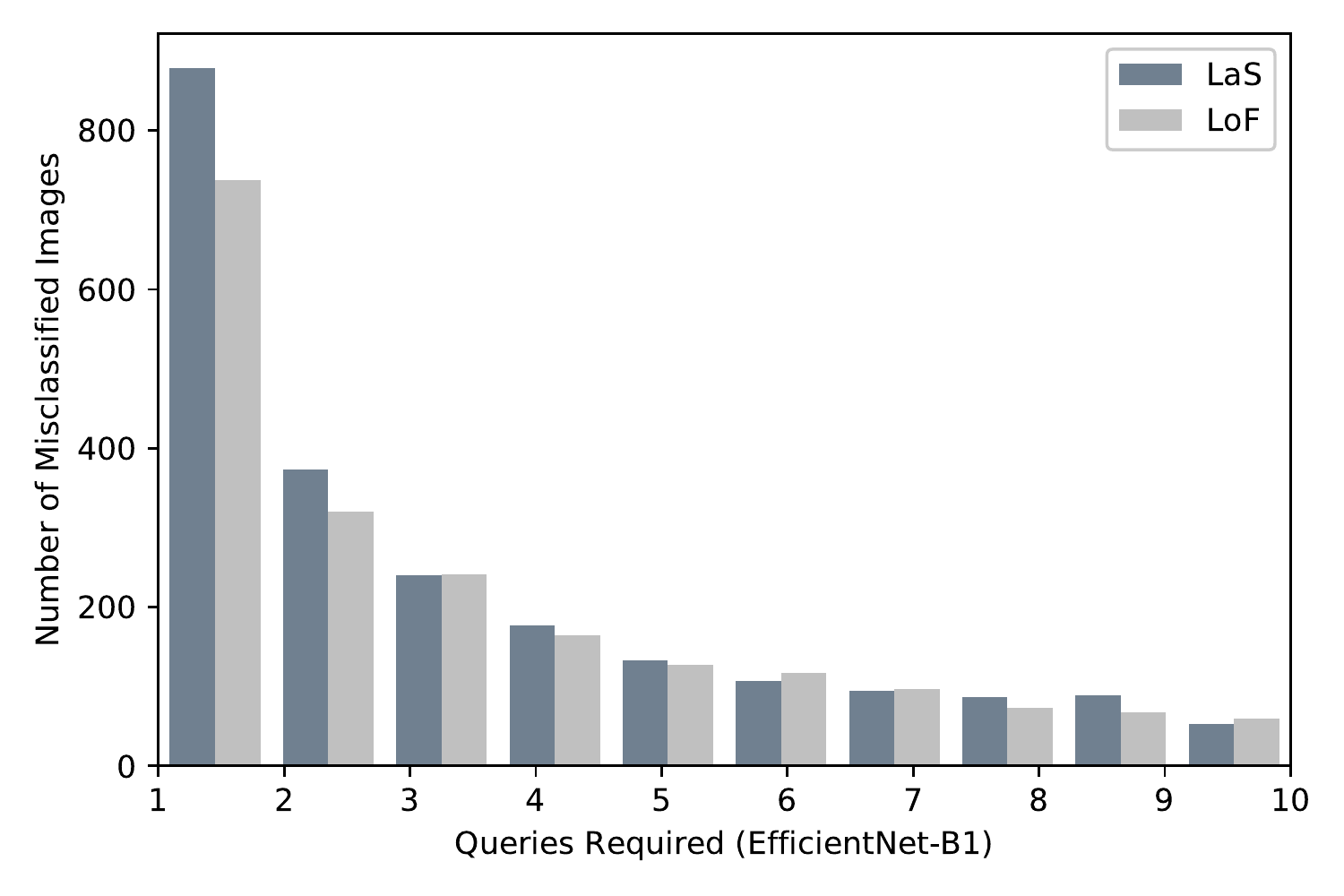}\par\medskip
\caption{Comparing the required number of queries to fool CNN models based on proposed approach (LaS), and LoF  \cite{Low_DCT1}.}
\label{fig:figur_query_dist}
\end{figure*}

\begin{figure}[h]
\centering
\includegraphics[height = 1.7in, width = 3.2in]{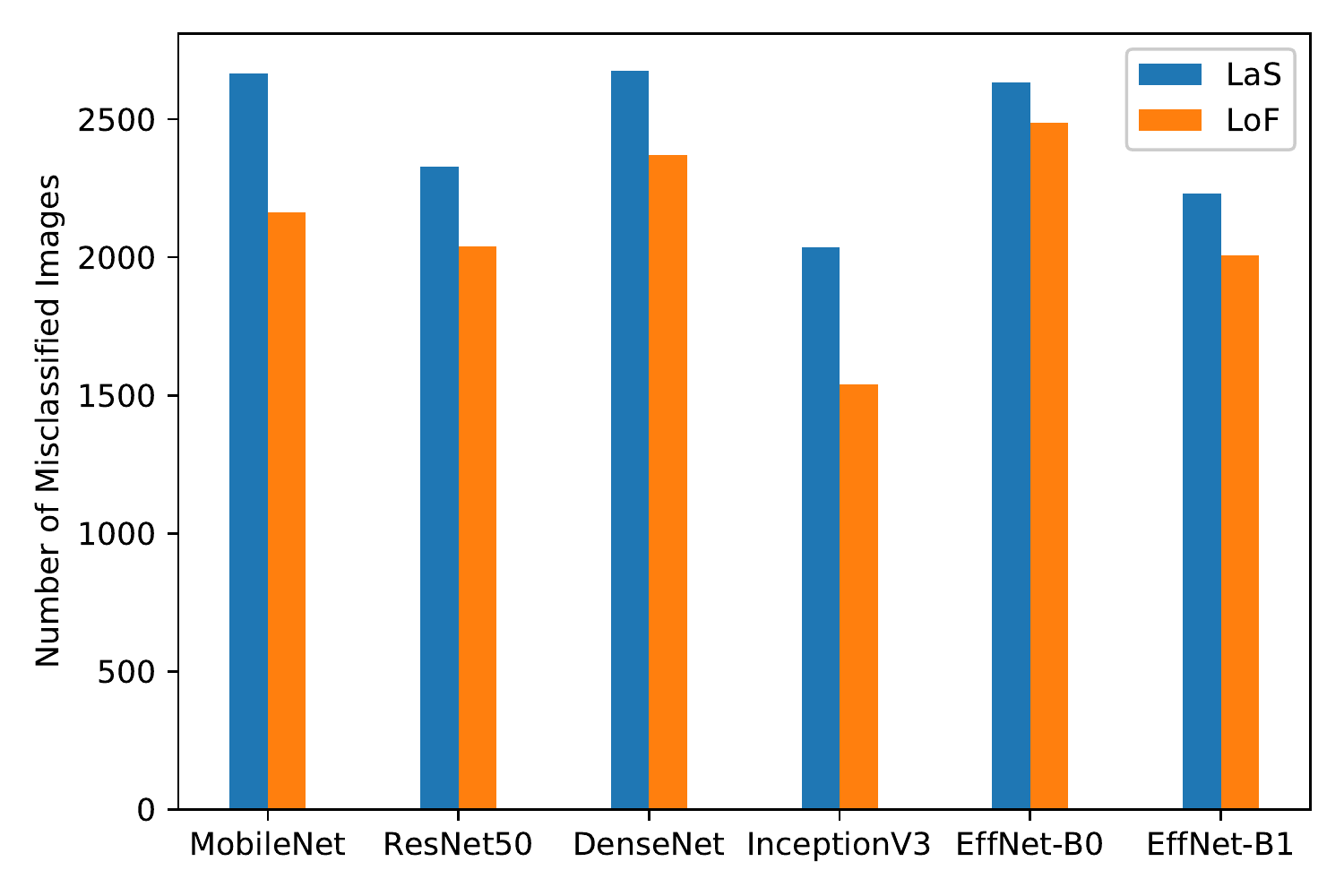}
\caption{Comparing number of misclassified samples for query less or equal to $10$ based on LaS and LoF \cite{Low_DCT1}.}
\label{fig:meanquery}
\end{figure}

In the adaptive black-box attack there is no prior information about the model's parameters and distribution of the training dataset, yet attacker can query the label of legitimate sample and corresponding perturbed sample. However, if the number of query to be increased, the system may identify a malicious activity. Obviously, an adversarial attack is more practical if it fools classifiers in a fewer number of queries. we designed a systematic experiment to evaluate the effectiveness of  adding perturbation on LaS components. Our results demonstrate that proposed approach requires fewer number of queries to fool image classifiers. In this experiment, six CNN models (EfficientNet-B0 and B1, ResNet50, InceptionV3, MobileNets, DenseNet121) were used. we trained all models with $50,000$ training samples of CIFAR 10 dataset. We used $10,000$ test samples of CIFAR-10 dataset that had never been used in training process to apply the attacks. We utilized DCT dictionary to transfer test samples into the frequency domain. We used a Gaussian noise with zero mean and variance $1$  to generate noise, and to have fair comparison with \cite{Low_DCT1}, we defined the MSE less than $0.001$ as a successful attack. We compared adding noise to $k=8$ LaS and LoF components.  In Fig. \ref{fig:figur_query_dist}, the histograms of required number of queries to successfully fool aforementioned CNN models are demonstrated. The distributions of successful attacks show that manipulating LaS components can fool the CNN models in a fewer number of queries. Figure \ref{fig:meanquery} shows the number of all misclassified images in query less or equal to $10$. In this experiment, we firstly evaluated the models' prediction for each legitimate sample. If a model predicted a legitimate sample wrongly, we put aside that sample and did not involve it to the experiment (because it was already misclassified). Hence, the number of misclassified images in Fig. \ref{fig:figur_query_dist} and \ref{fig:meanquery} are only due to the perturbation on samples.

\section{Case study: directed perturbation }
In this section, we propose a method for adding noise to the LaS components in order to fool the model into a specific direction. In the black-box approach,  the attacker can use some samples that have never been used for training stage. Then, the attacker can verify or find the input sample's label by observing the output of the objective model. In this section, we assume the attacker can have multiple samples of each class and their labels.  Suppose the available dataset is $\mathbf{X}=\ \left\{\mathbf{x}_i\right\}_{i=1}^{i=p}$ which contains $p$ samples and each sample belongs to one class out of $m$ available classes, i.e., $C(\mathbf{x}_i)\in\left\{c_j\right\}_{j=1}^{j=m}$. 
We map all samples of the dataset into the sparse domain via DCT dictionary $\mathbf{D}$. Doing so, $\mathbf{S}=\ \left\{\mathbf{s}_i\right\}_{i=1}^{i=p}$ would be obtained where $\mathbf{s}_i$ is the sparse representation of the $\mathbf{x}_i$. In the sparse domain, we keep the $k$ LaS components and force the rest of the components to zero. Then each sparse vector is normalized. Doing so, we would have

\begin{equation}
\hat{\mathbf{S}}=\{ \hat{\mathbf{s}}_i \}_{i=1}^{i=p} , {||{\hat{\mathbf{s}}_i}||}_0 = k , {||{\hat{\mathbf{s}}_i}||}_2 = 1
\label{eq:l4}
\end{equation}
where $||.||_0$ is the zero-norm of a vector which counts the number of non-zero elements of a vector.  Sparse vector ${\hat{\mathbf{s}}}_i$ contains information of the positions and normalized values of the $k$ largest elements of $\mathbf{s}_i$ which belong to class $C(s_i)$. Then for a given$\ \left({\hat{\mathbf{s}}}_i,\ C(s_i)\right)$, we find the most correlated sparse vector  $\left({\hat{\mathbf{s}}}_j,\ C(s_j)\neq C(s_i)\right)$. In other words, sparse vector ${\hat{\mathbf{s}}}_j$ is the closest sparse vector to the ${\hat{\mathbf{s}}}_i$, but they belong to different classes. We used the inner product of two vectors $\langle{\hat{\mathbf{s}}}_i,{\hat{\mathbf{s}}}_j\rangle$ to calculate the correlation. If we change the $k$ most important elements of ${\hat{\mathbf{s}}}_i$ with respect to the $k$ most important elements of ${\hat{\mathbf{s}}}_j$, some information and features of ${\hat{\mathbf{s}}}_j$ can be transferred into the ${\hat{\mathbf{s}}}_i$. If some nonzero elements of ${\hat{\mathbf{s}}}_i$ and ${\hat{\mathbf{s}}}_j$ have the same positions and close values, there is no need to change or manipulate them. Because they have common information and changing them cannot help for fooling classifier and may bring unnecessary perturbation in the pixel domain. To prevent this probable issue, we subtract these two vectors to obtain the difference $\mathbf{d}_{ij}$ as follows:
\begin{equation}
\mathbf{d}_{ij}=\ {\hat{\mathbf{s}}}_i-{\ \hat{\mathbf{s}}}_j
\label{eq:diff}
\end{equation}
Then, we subtract a multiplier of $\mathbf{d}_{ij}$ from the original sparse vector $\mathbf{s}_i$ to obtain sparse adversarial example ${\widetilde{\mathbf{s}}}_i$ as follows:

\begin{equation}
{\widetilde{\mathbf{s}}}_i=\ \mathbf{s}_i- \delta\mathbf{d}_{ij}
\label{eq:l6}
\end{equation}
where $\delta$ is a scalar number that controls the level of directed perturbation.  Then, we transfer back the adversarial sparse vector ${\widetilde{\mathbf{s}}}_i$ to the pixel domain via dictionary $\mathbf{D}$ as follows: 
\begin{equation}
{\widetilde{\mathbf{x}}}_i=\ \mathbf{D}{\widetilde{\mathbf{s}}}_i
\label{eq:l7}
\end{equation}
where ${\widetilde{\mathbf{x}}}_i$ is the adversarial example. Since the response of ML classifier for  $\mathbf{s}_j$ is $C(s_j)$, when we add the elements of ${\hat{\mathbf{s}}}_j$ to the ${\hat{\mathbf{s}}}_i$, the classifier may be fooled. By choosing $\delta$ and $k$ properly, ML classifiers can be fooled. Two scalar parameters $k$ and $\delta$ control the level of perturbation. When we increase these scalars, the level of perturbation in the pixel domain and misclassification rate would be increased accordingly.  Two error metrics to compare the adversarial image quality with the legitimate image are the Mean Square Error (MSE) and the Peak Signal to Noise Ratio (PSNR). The MSE yields the cumulative squared error between the adversarial and the legitimate image, whereas PSNR gives a measure of the peak error. The higher the value of PSNR, the higher the quality.
\begin{equation}
MSE=\frac{||{\mathbf{x}_i - {\widetilde{\mathbf{x}}}_i ||}_{2}^{2}}{N}
\label{eq:mse}
\end{equation}
\begin{equation}
PSNR=10\log_{10}{\left(\frac{h^2}{MSE}\right)}\label{eq:psnr}
\end{equation}
where $h$ is the maximum fluctuation in the input image data type. For example, since we normalized all image dataset to the range of $[0,1]$, input images' pixels fluctuate between zero and one, so $h=1$. Before investigating the relation between misclassification rate and quality metrics, we recall two important properties of the matrix-vector multiplications; first, the product of an orthonormal matrix by a vector does not change the norm-2 of that vector, and second, a scalar number can take out of the norm-2 of a vector. With respect to these two properties, since $||{\mathbf{x}_i - {\widetilde{\mathbf{x}}}_i ||}_{2}^{2}={||\delta \mathbf{D}\mathbf{d_{ij}}||}_{2}^{2}$ and due to the fact that the dictionary\ $\mathbf{D}$ is an orthonormal dictionary and the $\delta$ is a scalar value, $||{\mathbf{x}_i - {\widetilde{\mathbf{x}}}_i ||}_{2}^{2}=\delta^2{|| \mathbf{d_{ij}}||}_{2}^{2}$. Equation (\ref{eq:mse}) can be further simplified to obtain more straightforward relation between $\delta$ and $MSE$ or $PSNR$ in pixel domain as follows:
\begin{equation}
MSE=\frac{\delta^2}{N}{|| \mathbf{d_{ij}}||}_{2}^{2}=\frac{\delta^2}{N}{||{\hat{\mathbf{s}}_i}-{\hat{\mathbf{s}}_j}||}_{2}^{2}=\frac{2\delta^2}{N}
(1-\langle{\hat{\mathbf{s}}}_i,{\hat{\mathbf{s}}}_j\rangle)
\label{eq:mse2}
\end{equation}
where $\langle\cdot,\cdot\rangle$ is the inner product operation of two vectors. Since both ${\hat{\mathbf{s}}}_i$ and ${\hat{\mathbf{s}}}_j$ are normalized vectors, their inner product equals a number belongs to $\left[-1,1\right]$. Hence $MSE$ can be bounded $0\le MSE\le\frac{4\delta^2}{N}$. 
However, as we choose two most correlated sparse vectors, their inner product is usually greater than zero.  Hence, the upper bound of MSE may be smaller, i.e. $0\le MSE\le\frac{2\delta^2}{N}$. This inequality shows how adding perturbation in the sparse domain can be reflected in the perturbation in the pixel domain. The value of the $\delta$ directly affects the $MSE$. The order of sparsity, $k$, only has its effect on the inner product.\\
We applied the directed attack over the same six CNN models, and compared the effectiveness of adding noise to the LaS components against adding noise to the LoF components. In this experiment, we used multiple values for $\{k = 20, 30, 40\}$, and we fixed the value of $\delta$ in order to have $MSE \leq 0.001$. Table \ref{table:directed_attack} shows the results and superiority of manipulating LaS components.

\begin{table}
\caption{Comparing misclassification rates of directed attack over six CNN models based on proposed method (LaS) and recent method (LoF) \cite{Low_DCT1}.}
\label{table:directed_attack}
\begin{tabular}{|l|c|c|c|c|c|c|l|}
\hline
\multicolumn{1}{|c|}{{ }}   & \multicolumn{2}{c|}{\textbf{k=20}} & \multicolumn{2}{c|}{\textbf{k=30}}   & \multicolumn{2}{c|}{\textbf{k=40}}   \\ \cline{2-7} 
\multicolumn{1}{|c|}{\multirow{-2}{*}{{ Model}}} & \multicolumn{1}{l|}{{ \textbf{LaS}}} & \multicolumn{1}{l|}{\textbf{LoF} } & \multicolumn{1}{l|}{\textbf{LaS}} & \multicolumn{1}{l|}{\textbf{LoF}} & \multicolumn{1}{l|}{\textbf{LaS}} & \multicolumn{1}{l|}{\textbf{LoF}}  \\
\hline
MobileNets      & { 19.7}      & 19.3          & 22.3     & 21.5     & 23.6     & 22.9     \\ \hline
ResNet50        & { 21.9}      & 21.8          & 24.2      & 23.9     & 25.6    & 25.3      \\ \hline
DenseNet121     & { 20.0}      & 19.2         & 22.3       & 20.8     & 23.4    & 22.3     \\ \hline
InceptionV3     & { 16.4}      & 15.3         & 17.9      & 16.7      & 18.4    & 17.3      \\ \hline
Efficient -B0   & { 16.1}      & 15.6        & 18.8        & 17.7      & 20.2   & 19.6      \\ \hline
Efficient-B1    & 13.7        & 13.1        & 15.5        & 14.7     & 16.9     & 15.8        \\ \hline
\end{tabular}
\end{table}

\begin{figure}[!ht]
\centering
\includegraphics[height = 2.8in, width = 3.4in]{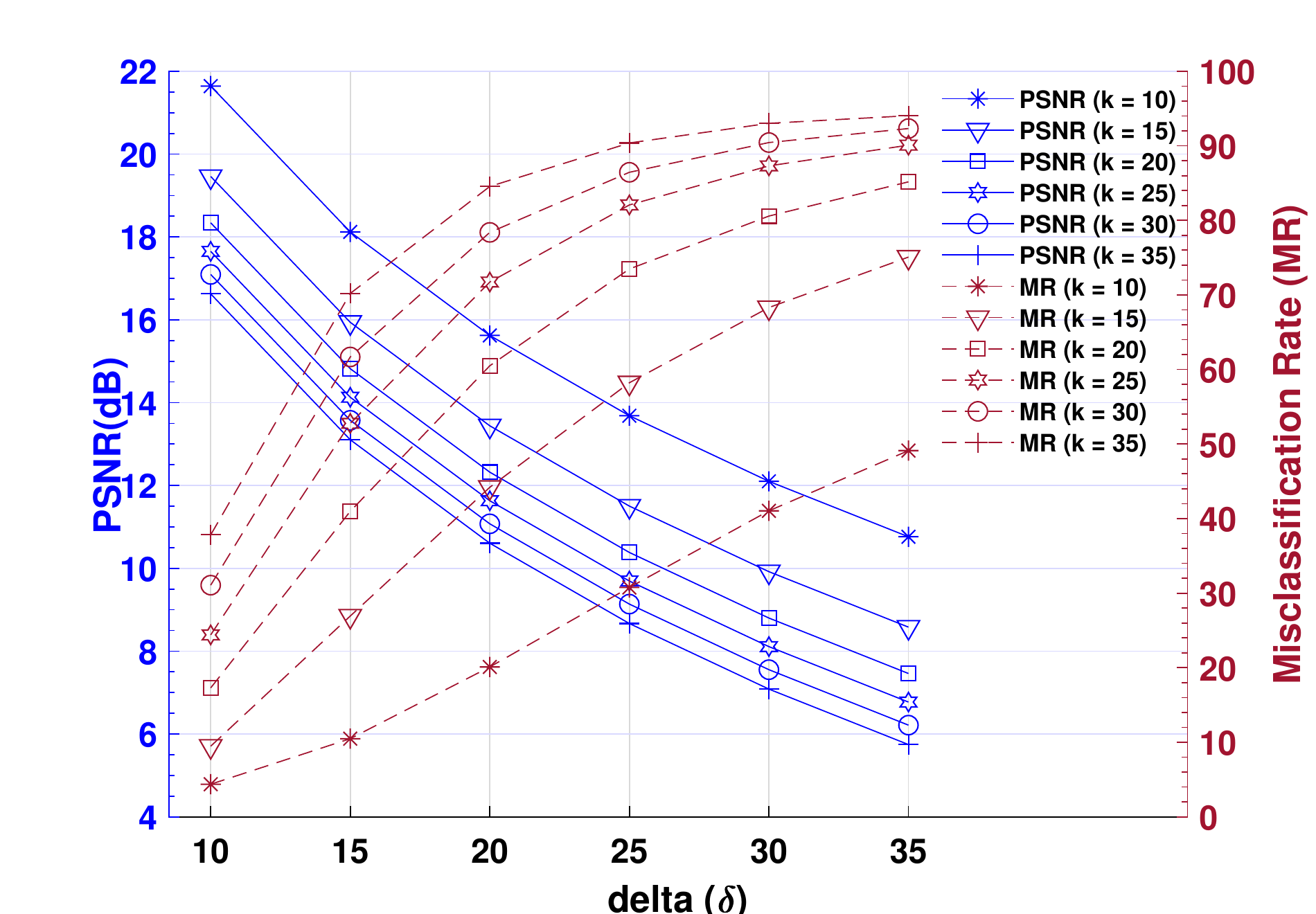}
\caption{Generating adversarial examples with different level of perturbation on LeNet.}
\label{fig:onLeNet}
\end{figure}
\begin{figure}[!ht]
\centering
\includegraphics[height = 2.8in, width = 3.4in]{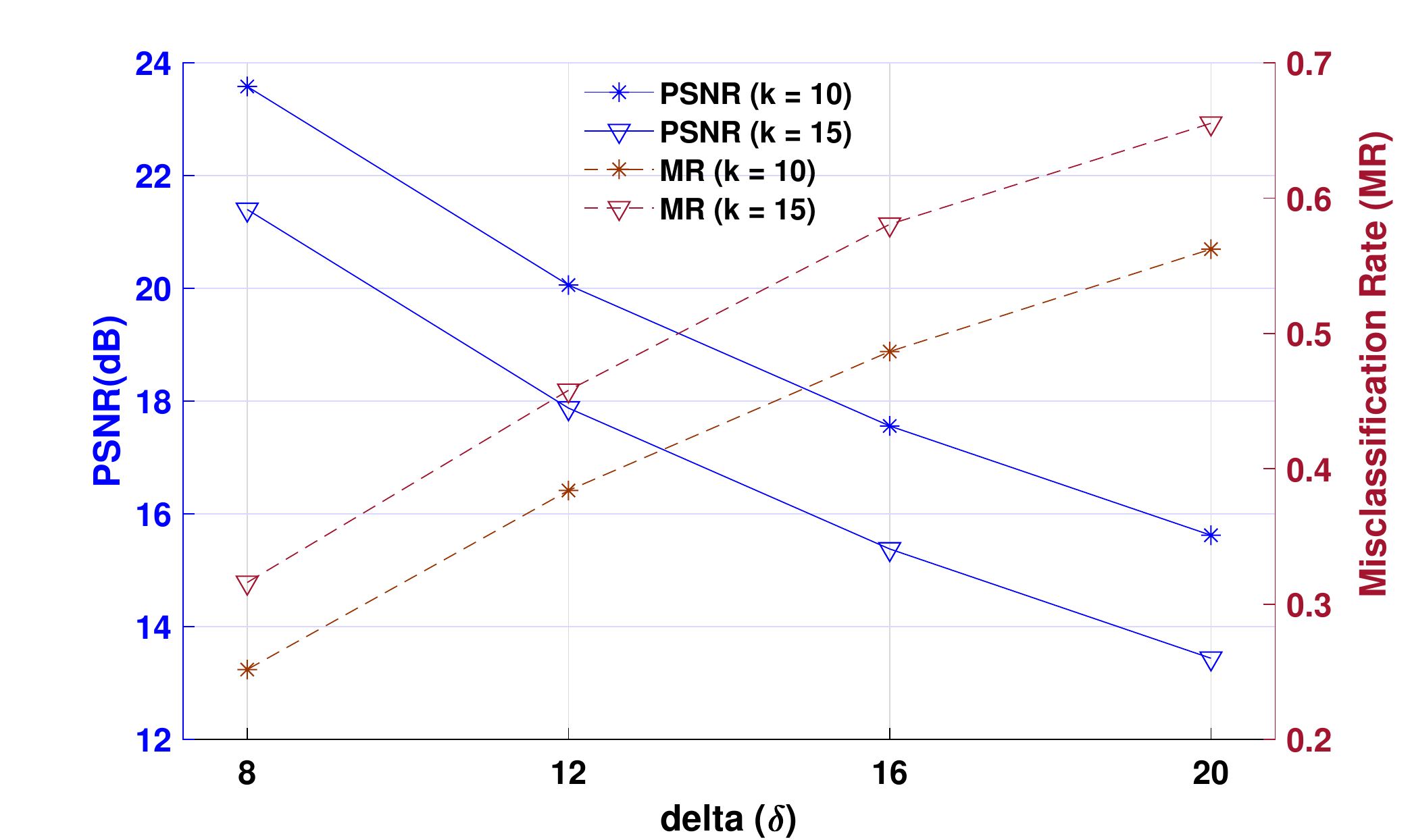}
\caption{Generating adversarial examples with different level of perturbation on SVM classifier.}
\label{fig:onSVM}
\end{figure}

As theoretically was discussed, changing $\delta$ can directly affect the perturbation level. To show this property, we trained the LeNet network \cite{LeNet} with $60,000$ training samples of MNIST dataset and achieved the accuracy of $98.2\%$ which means $1.8\%$ misclassification rate  over $10,000$ test samples. Then, we used the same test dataset and selected $6$ different values for the $\delta$ and $k$. It leads to running $36$ times, all combinations of $\delta$ and $k$ to generate corresponding perturbed test dataset. Then we input all these $36$ adversarial sets to the LeNet classifier to observe the response of the network. Figure \ref{fig:onLeNet} illustrates the effect of $\delta$ and $k$, PSNR, and misclassification rate of LeNet network. The left and right y-axes show the PSNR value the misclassification rate of each perturbed dataset, respectively. Solid blue lines show that PSNR decreases as delta value increases, and dash lines show that the misclassification rate increases as we increase the value of $\delta$. 
We also evaluated the effectiveness of our proposed attack on the SVM classifier. Due to the computational limitation, we only  used  $15000$ training and $3000$ test samples of MNIST dataset. After trying multiple kernels, the polynomial kernel was the best kernel to achieve the highest score for the classification. The misclassification rate of the trained SVM classifier on the benign test dataset was $5\%$. Then we generated  adversarial sets with different levels of perturbation. Figure \ref{fig:onSVM} shows that the SVM classifier is highly vulnerable to the proposed attack.  

\begin{figure}[!ht]
    \centering
    \includegraphics[height = 2.8in, width = 3.4in]{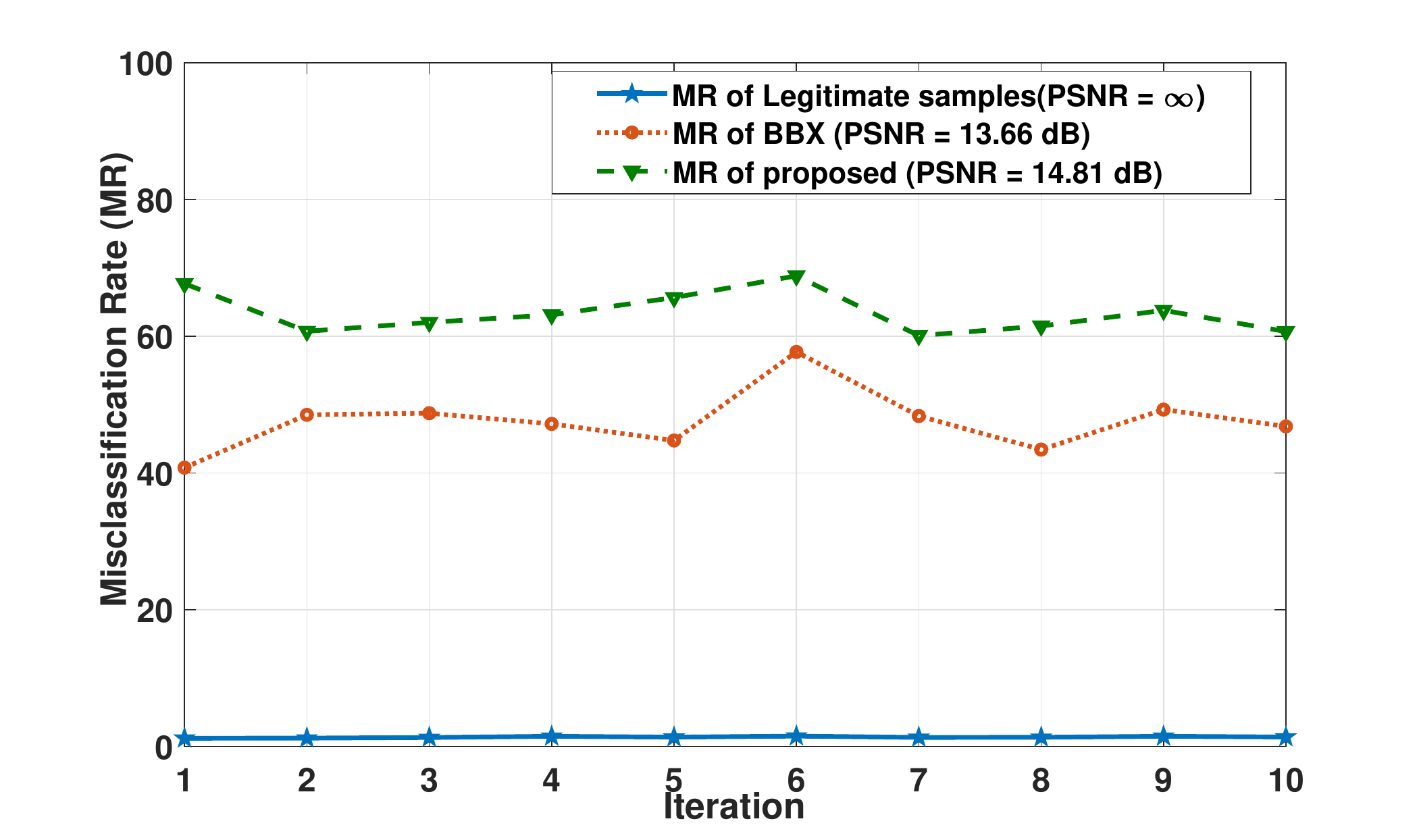}
    \caption{Comparing the misclassification rate of proposed method of perturbation and recent practical black-box (BBX) approach.  \cite{Papernot2017}. }
    \label{fig:bbx}
\end{figure}

We compared our approach with a recent work by Papernot et al \cite{Papernot2017} which is not based on frequency domain. We used the Cleverhans library \cite{cleverhans}, and to have a fair comparison, the same CNN and parameters were used. We trained the network $10$ times, and after each time the misclassification rate of the trained model on both adversarial sets was recorded. Figure \ref{fig:bbx} shows for $\delta =15$ and $k=20$,  our proposed adversarial examples have higher misclassification rate than that of the previous work, while our method has a higher PSNR which means less perceptible perturbation.     

\section{Attacking Google Cloud vision and YOLO}

\begin{figure}[h]
    \centering
    \includegraphics[height = 2.5in, width = 3.4in]{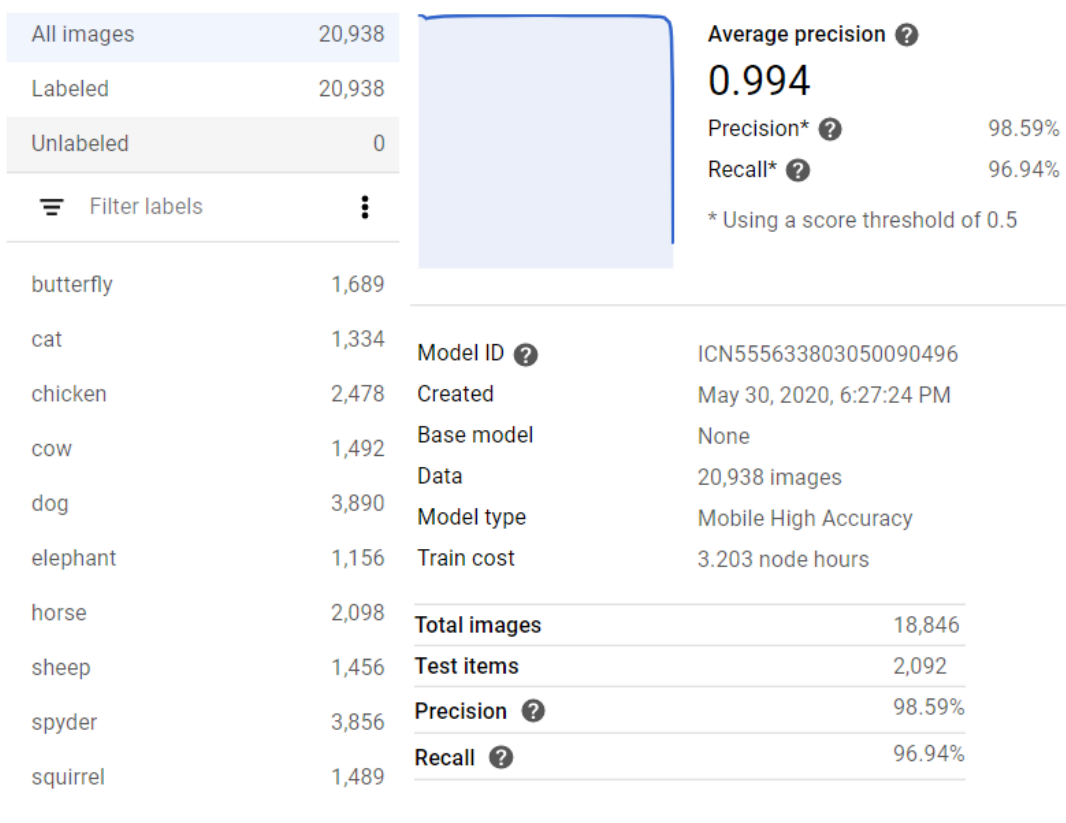}
    \caption{
    Information of dataset and trained model by Google Cloud Vision.} 
    \label{fig:googleModel}
\end{figure}

\begin{figure}[h]
    \centering
    \includegraphics[height = 2.5in, width = 3.4in]{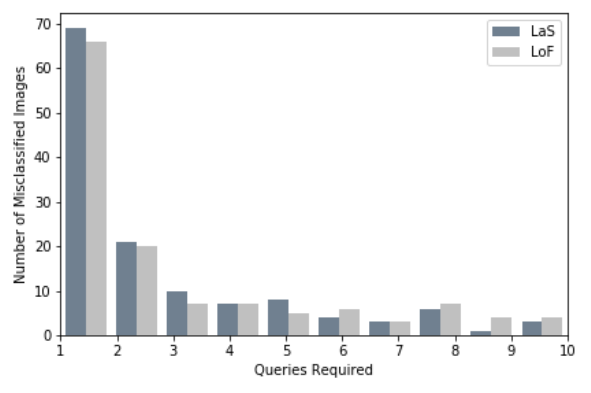}
    \caption{Comparing the required number of queries to fool a TFlite model trained by GoogleAPI based on proposed approach (LaS), and LoF  \cite{Low_DCT1}.} 
    \label{fig:googleModelcomparison}
\end{figure}

\begin{figure}[h]
    \centering
    \includegraphics[height = 3.2in, width = 3.4in]{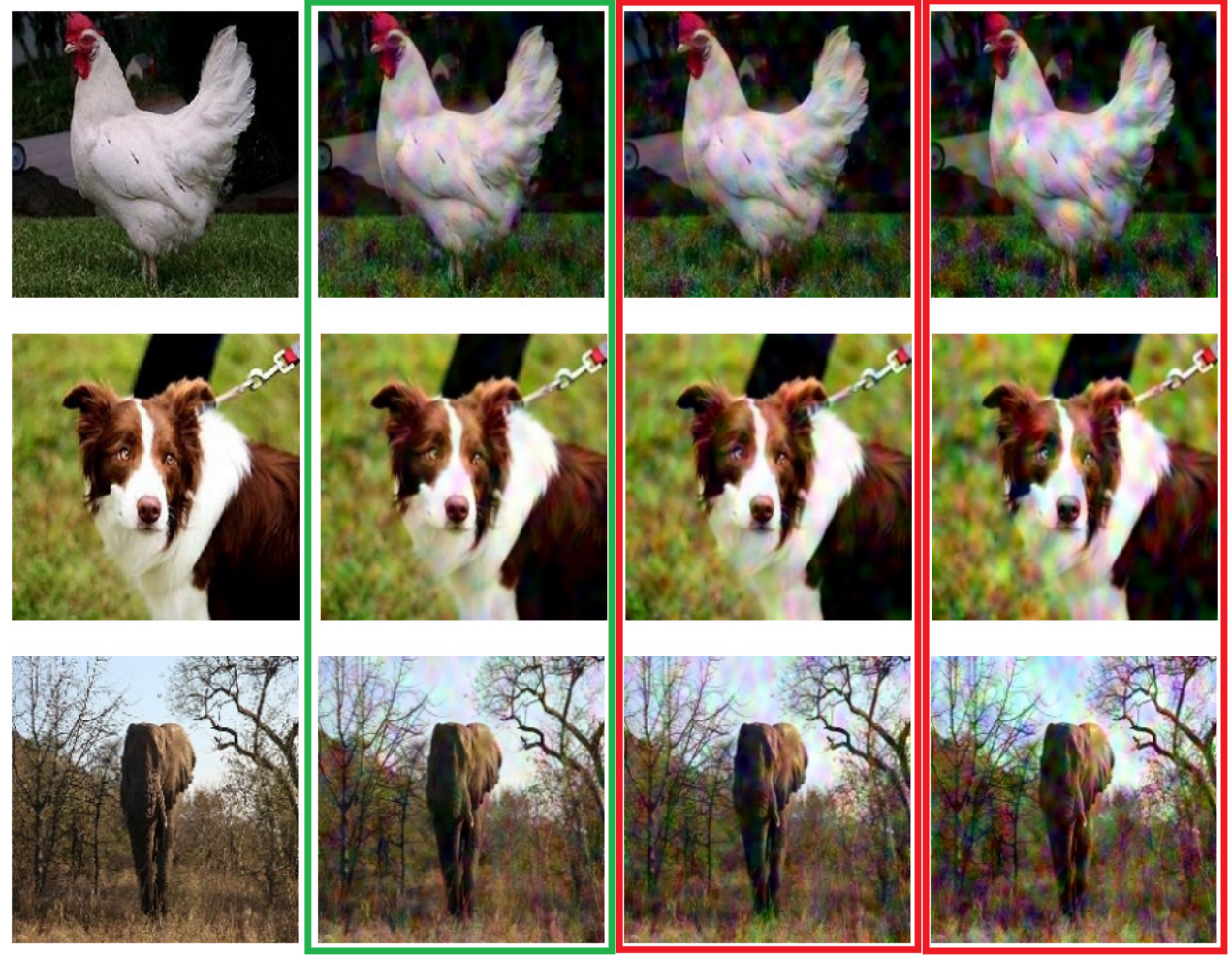}
    \caption{
    Samples of attacking Google Cloud Vision. The most left column corresponds to the legitimate image, and the other three columns are misclassified adversarial examples. 
    we supposed $MSE> 0.001$  as a failure and corresponding adversarial examples are bounded by red color boxes. We supposed $MSE=0.001$ as a success and corresponding examples are enclosed by a green color box.} 
    \label{fig:googleAPISamples}
\end{figure}

To evaluate the realistic threat of LaS components perturbation, we attacked a popular online machine learning service, Google Cloud Vision. The platform provides a TFLite version that can be deployed over Android operating systems \cite{TFLite}. We used a high-resolution dataset which contained $20938$ samples belong to $10$ animals ``spider, dog, cat, squirrel, sheep, butterfly, horse, elephant, cow, chicken" \cite{10Animal}. Figure \ref{fig:googleModel} shows the details of the trained model by Google Cloud Vision. To assess the effectiveness of our proposed attack, we downloaded its TFLite version. We randomly selected $500$ test samples and added perturbation based on LaS and LoF approaches. By adding limited noise to LaS components, $132$ samples out of $500$ samples were misclassified. Also, adding noise to LoF components led to $129$ misclassified samples. Figure \ref{fig:googleModelcomparison} shows the number of required queries to fool the TFlite model based on both methods. In addition, Fig. \ref{fig:googleAPISamples}  shows  three samples and corresponding adversarial examples for MSE values equal to $0.001$, $0.002$, and $0.005$. The first column shows the legitimate samples that are classified correctly by the classifier, the second column from the left which closed by a green box, belongs to the adversarial examples with $MSE=0.001$, the other two columns with red boxes related to the adversarial examples with $MSE=0.002$ and $0.005$. As defined in \cite{Low_DCT1}, we set the threshold of $MSE\leq 0.001$ 
as a successful attack.

\begin{figure}[!ht]
    \centering
    \includegraphics[height = 2.4in, width = 3.4in]{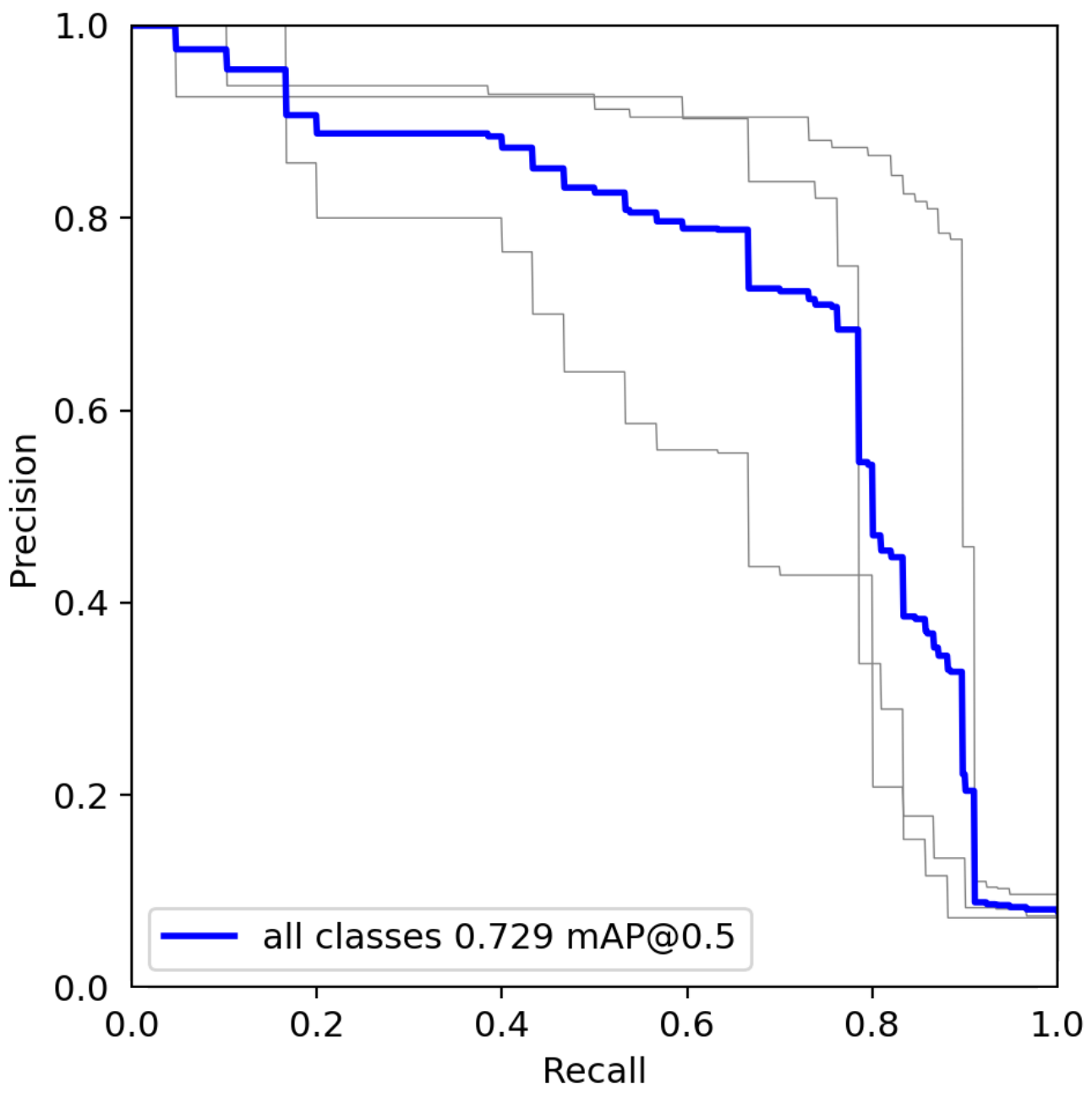}
    \caption{Performance of YOLOv5 over skin lesion dataset (ISIC-2017).  }
    \label{fig:object_detection_acc}
\end{figure}

\begin{figure}[ht]
    \centering
    \includegraphics[height = 3.2in, width = 3.2in]{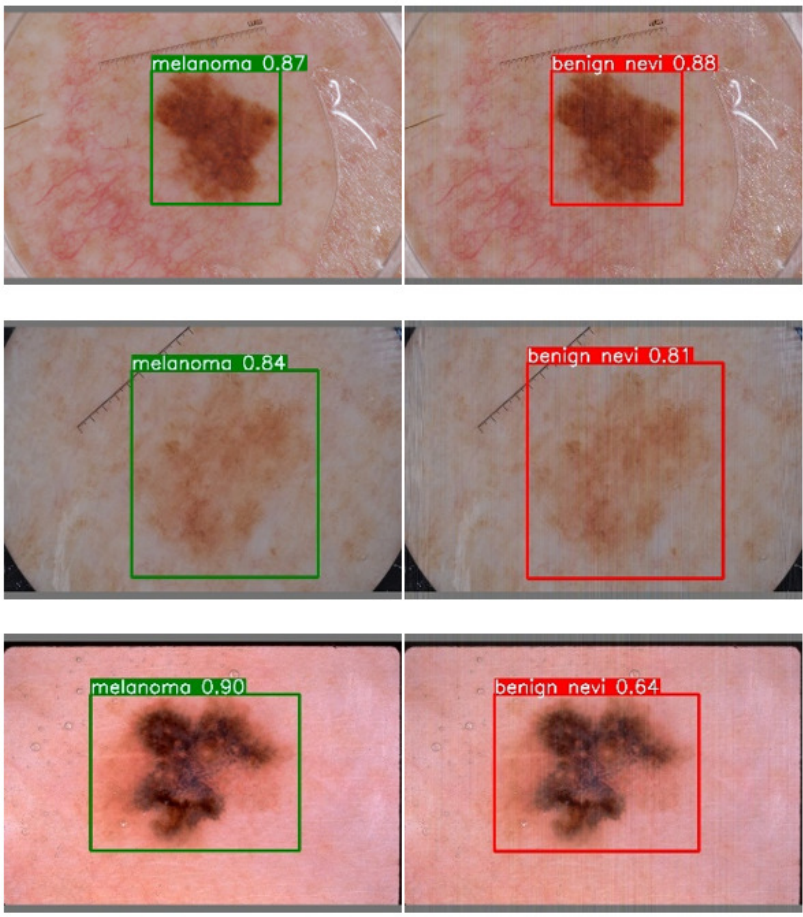}
    \caption{
    Samples of attacking the object detection algorithm (YOLOv5). The left column corresponds to the legitimate images that have been correctly detected and classified, and the right column corresponds to the misclassified objects.} 
    \label{fig:object_detection}
\end{figure}

In addition, we applied our attack over an object detection algorithm. Object detection has been widely used by autonomous vehicles and biomedical devices. One of the fastest and most accurate object detection algorithms is YOLOv5 \cite{yolov5-ref}. YOLOv5 is a one-stage algorithm that  implements classification and regression tasks in a single step. Object detection algorithms implement two tasks, detection and classification. In certain sensitive applications, if the model fails to detect the object correctly or predict the label wrongly, it may cause irreversible consequences. In this experiment,  we used \textit{International Skin Imaging Collaboration} (ISIC)-2017 skin lesion dataset that contains $2000$ training samples, $150$ validation samples, and $600$ test samples belong to three skin lesion classes: \textit{melanoma,  nevus, and seborrheic keratosis}. We resized the input samples into $640$x$640$ pixels and set two parameters as Intersection over Union (IoU) to $0.50$ and confidence threshold to $0.25$. We trained the model and evaluated its performance over $600$ test samples. Figure \ref{fig:object_detection_acc} shows the performance of trained model over test dataset. Precision is a metric that measures how accurate is the predictions, while recall measures how good the model finds all the positive cases. IoU measures the overlap between predicted box around the object with the ground truth. The model achieved mean Average Precision (mAP) equal to $0.72$ over three classes. In next step, we randomly selected some test samples that had never been used in training process to add perturbation and observe the model response. Our results show that by adding limited noise to the LaS components, this model predicts wrong labels with high confidence scores. In Fig. \ref{fig:object_detection}, we only showed few adversarial examples that had been misclassified. However, there were adversarial samples that model could not detect any object. In this experiment, we set $MSE\leq 0.001$ to generate adversarial examples. We released our code, the TFlite model trained by Google Cloud Vision, trained object detection model, and the annotation files of ISIC-2017 dataset publicly for reproducibility \cite{github}.

\section{Conclusion}
\label{Conclusion}
In this work, we proposed a new approach for generating adversarial examples in the sparse domain. We show LaS components are different from LoF components, and they belong to all frequency bands (low, middle, or high). We proposed a hypothesis that LaS components affect the decision boundaries of CNN models much more than LoF components. This hypothesis was the key to build our proposed adversarial method. We designed a systematic experiment to support this hypothesis. By running experiments over six advanced CNN models, we empirically verified that LaS components affect decision boundaries of CNN models more than LoF components. Then we added a limited noise to the LaS components to generate our proposed adversarial example. We evaluated the response of six advanced CNN models against our adversarial examples and compared it with recent work. Our results over MNIST and CIFAR-10 datasets unanimously support this hypothesis that adversarial examples generated based on manipulating LaS components, can fool the CNN models in much fewer number of queries than that of the LoF approach.  We also implemented our experiments over Animal and skin lesion ISIC-2017 datasets to evaluate Google Cloud Vision API and YOLO algorithm. Results show the effectiveness of our proposed method to fool aforementioned models. By introducing the potential threat within this type of attack, an appropriate defense mechanism can be investigated in the future. Moreover, we used DCT dictionary to transfer images into the sparse domain, however, there are many other ways to transfer an image into a sparse domain other than the DCT domain that can be further investigated.

\begin{IEEEbiography}[{\includegraphics[width=1in,height=1.25in,clip,keepaspectratio]{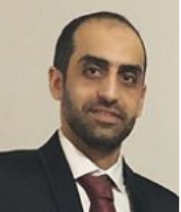}}]{Hadi Zanddizari}   is currently a research assistant  in the Department of Electrical Engineering at the University of South Florida. His research interests include deep learning, object detection, semantic segmentation, cybersecurity, and data privacy.
\end{IEEEbiography}

\begin{IEEEbiography}[{\includegraphics[width=1in,height=1.25in,clip,keepaspectratio]{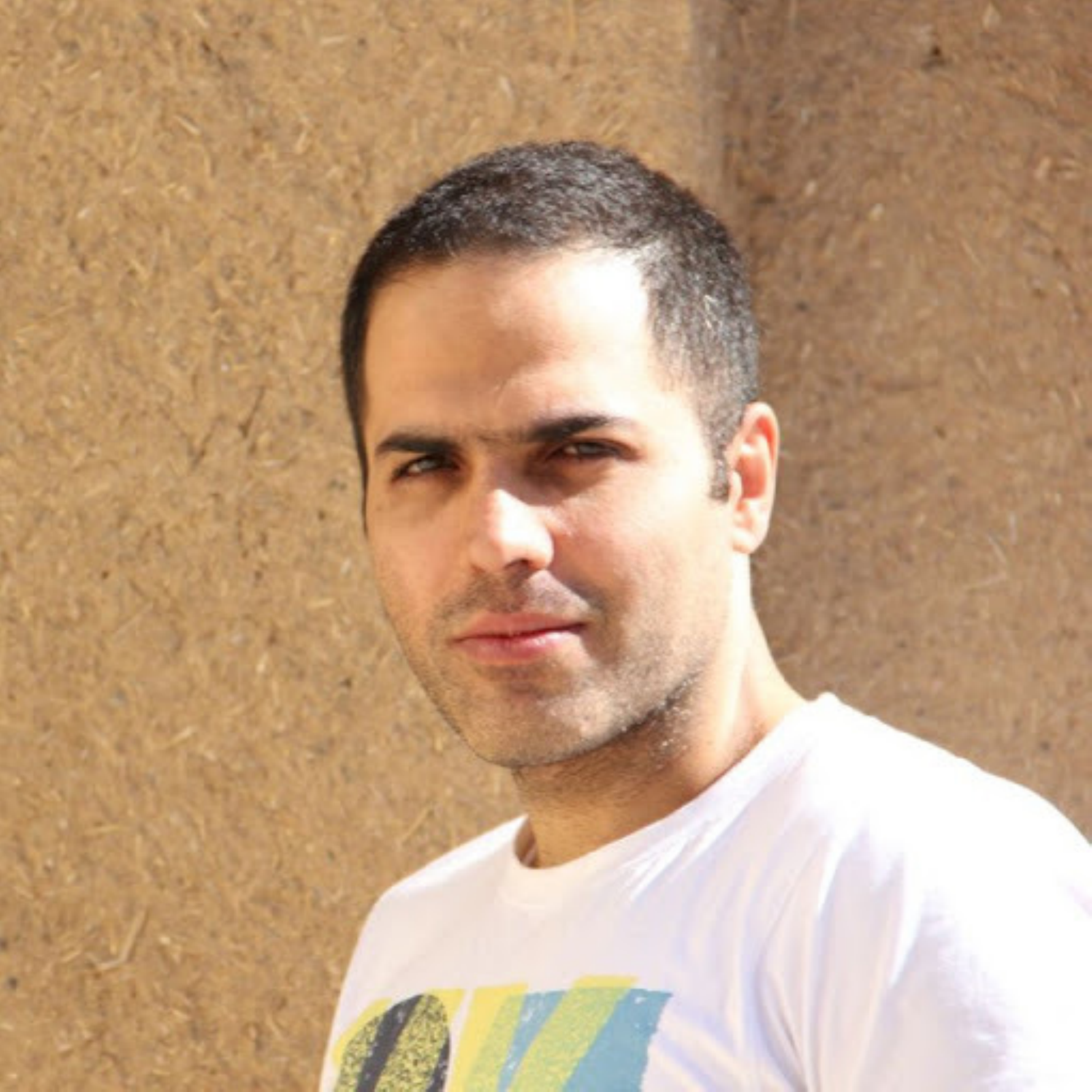}}]{Behnam Zeinali} received his MSc in Electrical Engineering from Iran University of Science and Technology, Iran, in 2013. From 2013 to 2019 he has worked in the industry as a programmer, researcher, and developer in the field of AI. Currently, he is working towards a Ph.D. degree from the University of South Florida. His research focuses are on the machine and deep learning, computer vision, and mobile application programming.
\end{IEEEbiography}

\begin{IEEEbiography}[{\includegraphics[width=1in,height=1.25in,clip,keepaspectratio]{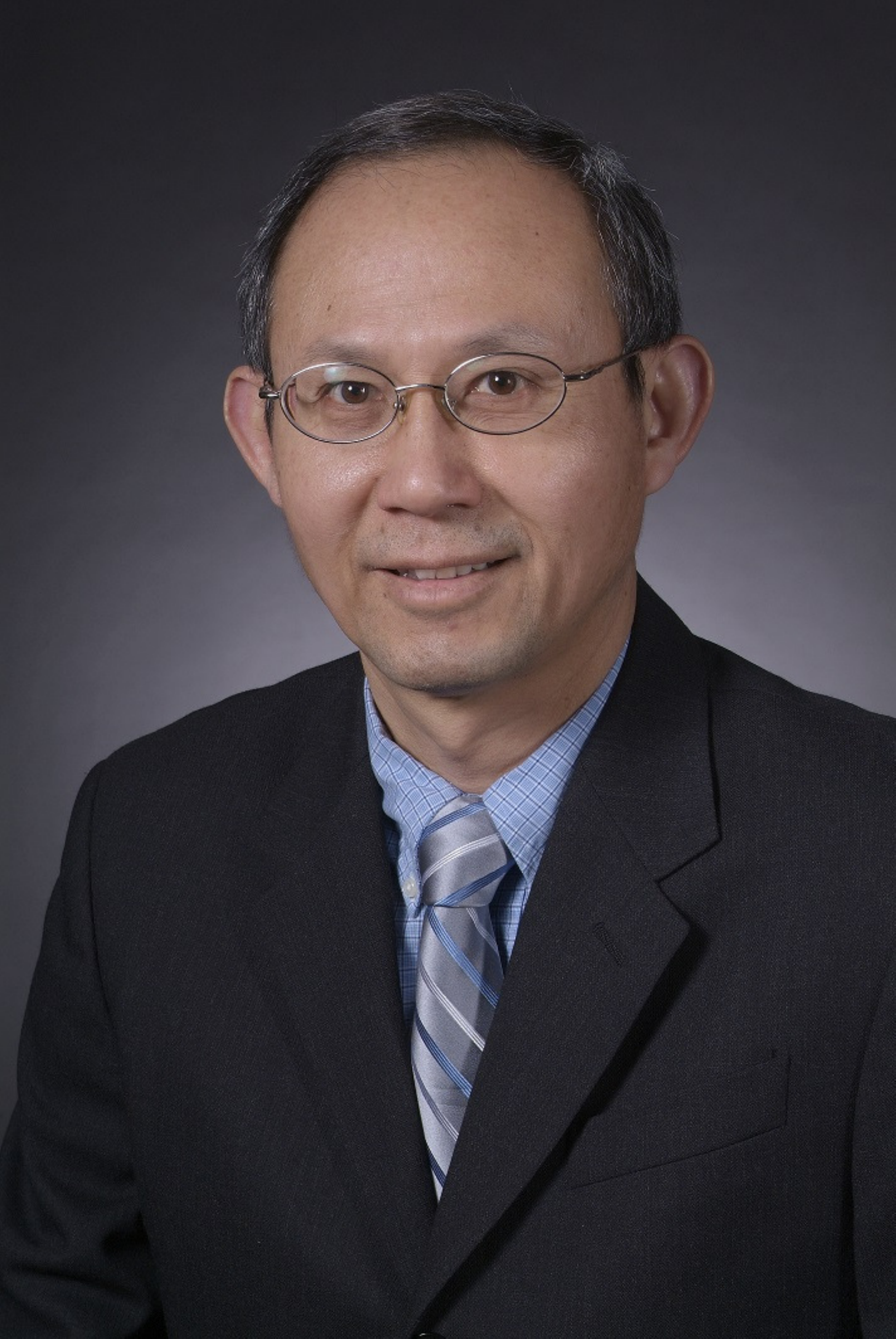}}]{J. Morris Chang} is a professor in the Department of Electrical Engineering at the University of South Florida. He received the Ph.D. degree from the North Carolina State University. His past industrial experiences include positions at Texas Instruments, Microelectronic Center of North Carolina and AT \& T Bell Labs. He received the University Excellence in Teaching Award at Illinois Institute of Technology in 1999. He was inducted into the NC State University ECE Alumni Hall of Fame in 2019. His research interests include: cyber security and data privacy, machine learning, and mobile computing. He is a handling editor of Journal of Microprocessors and Microsystems and an editor of IEEE
IT Professional. He is a senior member of IEEE.
\end{IEEEbiography}

\end{document}